\pdfoutput=1

\documentclass[11pt]{article}

\usepackage[final]{acl}

\usepackage{array}
\usepackage{enumitem}
\usepackage{times}
\usepackage{latexsym}
\usepackage{amsmath}
\usepackage{cleveref}
\usepackage{booktabs}
\usepackage{multirow}
\usepackage{pifont}
\usepackage{hyperref}       
\usepackage{url}
\usepackage{amsfonts}
\usepackage{microtype}
\usepackage{graphicx}
\usepackage{colortbl}
\usepackage{amssymb}
\usepackage{tcolorbox}
\usepackage{tabularx}     
\usepackage{caption}      
\usepackage[utf8]{inputenc}
\usepackage{longtable}

\usepackage{booktabs} 
\usepackage{pifont}   
\usepackage{booktabs}   
\usepackage{pifont}     
\usepackage{tabularx}   

\definecolor{Maroon}{rgb}{0.5, 0, 0}  
\definecolor{OliveGreen}{rgb}{0.33, 0.42, 0.18}  
\definecolor{Violet}{cmyk}{0.25, 0.5, 0, 0}
\usepackage[T1]{fontenc}

\usepackage[utf8]{inputenc}

\usepackage{microtype}

\usepackage{inconsolata}

\usepackage{graphicx}
\newcommand{\bcmark}{\color{Violet}{\ding{51}}}%
\newcommand{\notcheckmark}{\textcolor{black}{\bcmark\kern-1.1ex\raisebox{.7ex}{\rotatebox[origin=c]{125}{--}}}\color{black}}

\title{Automating Hardware Design and Verification from Architectural Papers via a Neural-Symbolic Graph Framework}

\author{
    \textbf{Haoyue Yang\textsuperscript{1,}\footnotemark[1]},
    \textbf{Xuanle Zhao\textsuperscript{1,}\footnotemark[1]},
    \textbf{Yujie Liu\textsuperscript{1}}, 
    \textbf{Zhuojun Zou\textsuperscript{1}},
    \\
    \textbf{Kailin Lyu\textsuperscript{1}},
    \textbf{Changchun Zhou \textsuperscript{2,}\footnotemark[2]}, 
    \textbf{Yao Zhu\textsuperscript{3,}\footnotemark[2]}, 
    \textbf{Jie Hao\textsuperscript{1,}\footnotemark[2]}
    \\
\textbf{\textsuperscript{1}} Institute of Automation, Chinese Academy of Sciences, Beijing, China
\\
\textbf{\textsuperscript{2}} Peking University, Beijing, China
\textbf{\textsuperscript{3}} Zhejiang University, Zhejiang, China 
\\
\texttt{yanghaoyue2024@ia.ac.cn, ee\_zhuy@zju.edu.cn}
}

\begin{document}
\maketitle
\begin{abstract}
The reproduction of hardware architectures from academic papers remains a significant challenge due to the lack of publicly available source code and the complexity of hardware description languages (HDLs). To this end, we propose \textbf{ArchCraft}, a Framework that converts abstract architectural descriptions from academic papers into synthesizable Verilog projects with register-transfer level (RTL) verification. ArchCraft introduces a structured workflow, which uses formal graphs to capture the Architectural Blueprint and symbols to define the Functional Specification, translating unstructured academic papers into verifiable, hardware-aware designs. The framework then generates RTL and testbench (TB) code decoupled via these symbols to facilitate verification and debugging, ultimately reporting the circuit's Power, Area, and Performance (PPA). Moreover, we propose the first benchmark, \textbf{ArchSynthBench}, for synthesizing hardware from architectural descriptions, with a complete set of evaluation indicators, 50 project-level circuits, and around 600 circuit blocks. We systematically assess ArchCraft on ArchSynthBench, where the experiment results demonstrate the superiority of our proposed method, surpassing direct generation methods and the VerilogCoder framework in both paper understanding and code completion. Furthermore, evaluation and physical implementation of the generated executable RTL code show that these implementations meet all timing constraints without violations, and their performance metrics are consistent with those reported in the original papers.
\end{abstract}

\section{Introduction}

\begin{figure}[t]
    \centering
    \includegraphics[width=0.45\textwidth]{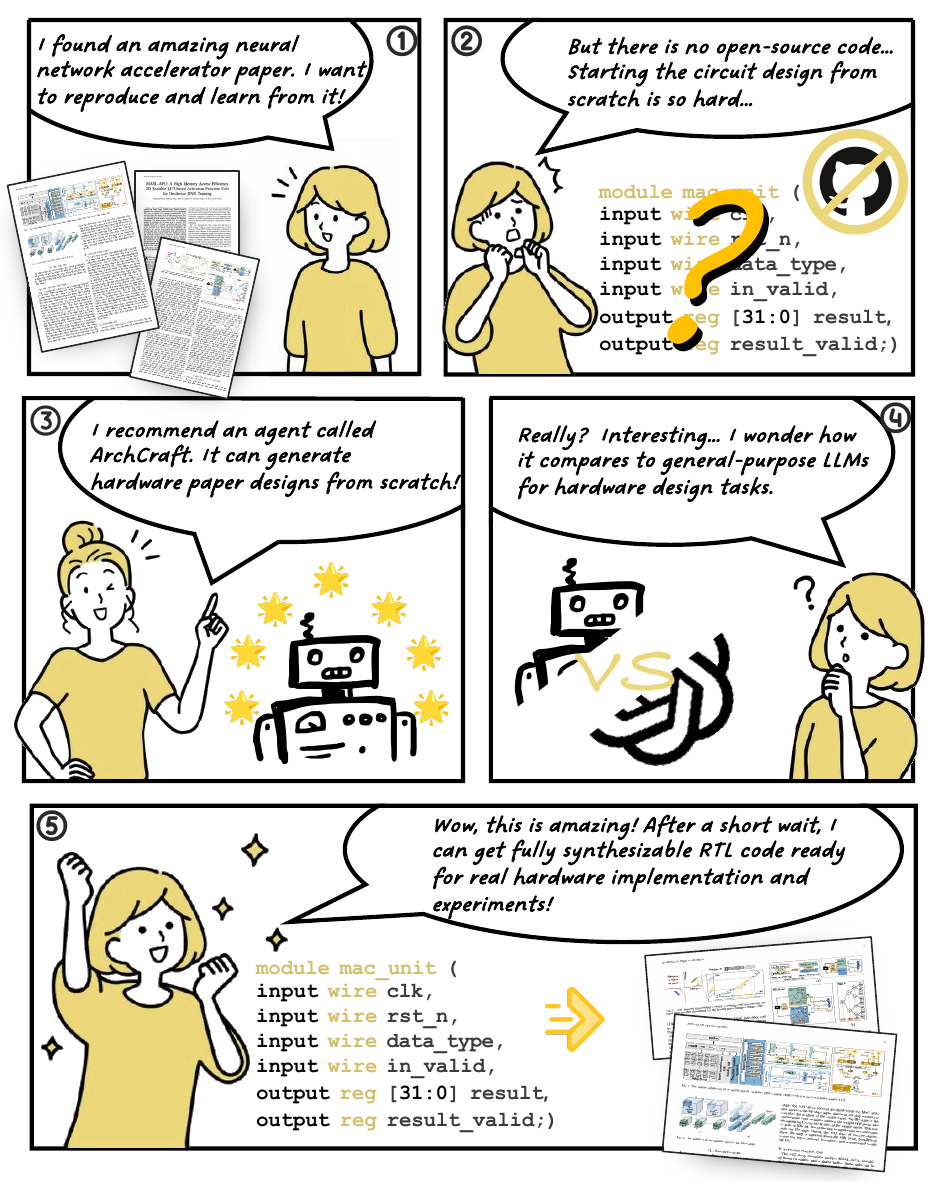}
    \vspace{-5pt}
    \caption{A real-world example. Archcraft assists in automating the process of transforming hardware design concepts into fully synthesizable RTL code, reducing manual intervention and accelerating the design process.}
    \label{fig:figure_1}
    \vspace{-15pt}
\end{figure}

“\textit{This paper on a novel neural network accelerator is excellent},” your hardware director mused, “\textit{but its implementation is closed-source. We must reproduce it to build our next-generation products upon it}.” This demand to reproduce synthesizable hardware designs from complex academic papers increasingly points to a fundamental challenge as shown in Figure \ref{fig:figure_1}: how can we design an automated system to seamlessly convert unstructured academic text, a 30-page PDF, into rigorous, synthesizable, and functionally correct code \cite{qian2023chatdev, seo2504paper2code, zhao2025autoreproduce, zhao2025chartcoder, zhang2023repocoder}? 

Let us deconstruct the workflow of a human expert performing this task. First, the engineer thoroughly reads the paper to construct a high-level architectural blueprint to identify key modules, their hierarchy, and the data flow connections between them. Next, for each module, the engineer defines its precise functional specification, including interfaces like inputs and outputs, internal states, and, critically, concurrent behaviors and timing logic. Finally, the engineer translates this structured, formalized understanding into RTL code and a corresponding TB. This process is, in essence, a complex, multi-step, domain-knowledge-dependent workflow.


Currently, approaches for any-to-any generative tasks typically fall into two paradigms: Implicit Neural Modeling and Agent Approaches. Implicit Neural Modeling approaches directly learn a neural representation from mass training data \cite{liu2024rtlcoder, liu2025deeprtl2, chang2024data, thakur2024verigen, wu2025itertl}. While Implicit Neural Modeling approaches show advantages in specific functional tasks and design space exploration, their extensibility is constrained by the scope of the training data. Consequently, it fails to understand the project-level architectural blueprint. The second, Agent Approaches \cite{zhao2025mage, ho2025verilogcoder, wang2025rtlsquad}, however, existing implementations often manifest as a Flat or Naive Agent paradigm. They fail to inherit the expert's actual mental model; instead, many rely on waveforms or TB as primary inputs. This is not true generation from scratch and fundamentally sidesteps the core challenge of translating high-level textual specifications into functional hardware logic.

This motivates us to explore an evolution of this paradigm: how can we design a "Structured Agentic Framework"? It should inherit the autonomy of the workflow while simultaneously embedding the human expert's mental model, enabling it to become more intelligent and reliable without training. Our core insight is: \textit{a successful hardware generation framework must replace the LLM's general-purpose reasoning with a domain-specific, formalized intermediate representation}. Based on this, we propose ArchCraft, a novel, training-free, structured agentic framework. The core of ArchCraft is the Graphing Agent, which systematically constructs the first part of the human mental model, the architectural blueprint, through a rigorous process. Its execution process involves defining the graph's scope and theme, identifying its nodes and static links, planning its directionality and internal node attributes, and finally, adding global constraints. Moreover, we introduce the Symboling stage, which systematically translates the node logic and attributes defined during the Graphing stage into a rigorous symbolic representation. This symbolic blueprint serves as a precise intermediate language, greatly facilitating the subsequent coding stage and constraining the LLM to adhere to correct hardware logic without any domain-specific model fine-tuning. Subsequent Coding and Evaluating agents then generate the RTL, TB, and perform functional verification based on this foundation. 



Furthermore, we constructed ArchSynthBench, the first benchmark specifically designed for synthesizing hardware from academic papers describing closed-source accelerators. The first key feature of ArchSynthBench is that it comprises 50 project-level circuits and around 600 circuit blocks. The second key feature of this benchmark is its complete set of evaluation indicators, which are meticulously structured into two levels, eight dimensions as detailed in Section \ref{subsec:evaluation}: evaluating architectural understanding and implementation completeness. Experiments on this benchmark show that our structured agentic framework achieves superior performance, surpassing LLMs with direct generation and agent frameworks like VerilogCoder. Benefiting from the Graph-based process, our paper-level assessment reaches a score of 86.17. Meanwhile, the symbolic constraints from Symboling ensure code-level completeness, achieving 81.04. These results demonstrate the framework's effectiveness on the novel task of zero-shot generation of synthesizable hardware.
In summary, the main contributions are as follows.

\begin{itemize}[leftmargin=*]
\vspace{-2mm}
\item We propose \textbf{ArchCraft}, a novel neural-symbolic graph-based framework that enables the complex task of transforming unstructured academic papers into synthesizable, project-level hardware implementations.
\vspace{-2mm}
\item Our proposed symbolic representation of functions and interfaces bridges the gap between RTL documentation and code, enabling the generation of non-open-source HDL code from models.
\vspace{-2mm}
\item We construct a benchmark, \textbf{ArchSynthBench}, to rigorously evaluate the replication accuracy and implementation quality of hardware synthesis. Our evaluation results show that previous methods struggle to reproduce functionally correct code, even for circuits.
\vspace{-2mm}
\item We further implement and evaluate the hardware design at the circuit level. The results show that the PPA of the synthesized implementation is consistent with that reported in the paper.
\end{itemize}

\section{Related Work}

\subsection{LLMs for Research}
LLM-agents are increasingly central to scientific discovery, from ideation to execution. Frameworks decompose research into autonomous modules for tasks like mining, experimentation, and writing \cite{schmidgall2025agentrxiv, schmidgall2025agent, ghafarollahi2025sciagents, baek2024researchagent, cui2025lumi, zhao2025vincicoder}, enhancing efficiency and scalability. For reproducibility, AutoReproduce \cite{zhao2025autoreproduce} uses a paper lineage algorithm to extract implicit knowledge, automating AI experiment reproduction with demonstrated superior performance. Collectively, these works show the potential of multi-agent LLM frameworks to transform the scientific process.

\subsection{LLMs for Hardware Design}
LLM application in hardware design is growing. Foundational datasets underpin this progress, such as PyraNet \cite{nadimi2024pyranet}, CodeV \cite{zhao2024codev}, and MG-Verilog \cite{zheng2024mgVerilog}. Methodologies include direct generation and agent-based frameworks. Direct generation includes Chip-Chat \cite{blocklove2023chip} for interactive co-design, and structured prompting methods like HiVeGen \cite{tang2024hivegen}, ROME \cite{nakkab2024rome}, SynC-LLM \cite{liu2025sync} and VeriMind \cite{nadimi2025verimind} (using chain-of-thought) to improve correctness and precision. Agent-based frameworks include VerilogCoder \cite{ho2025verilogcoder}, which integrates planning, checking, and simulation, and RTLSquad \cite{wang2025rtlsquad}, a collaborative agent architecture for integrated design, implementation, and verification, advancing hardware automation.

\renewcommand{\dblfloatpagefraction}{.6}
\begin{figure*}[htbp]
    \centering
    \includegraphics[width=\textwidth]{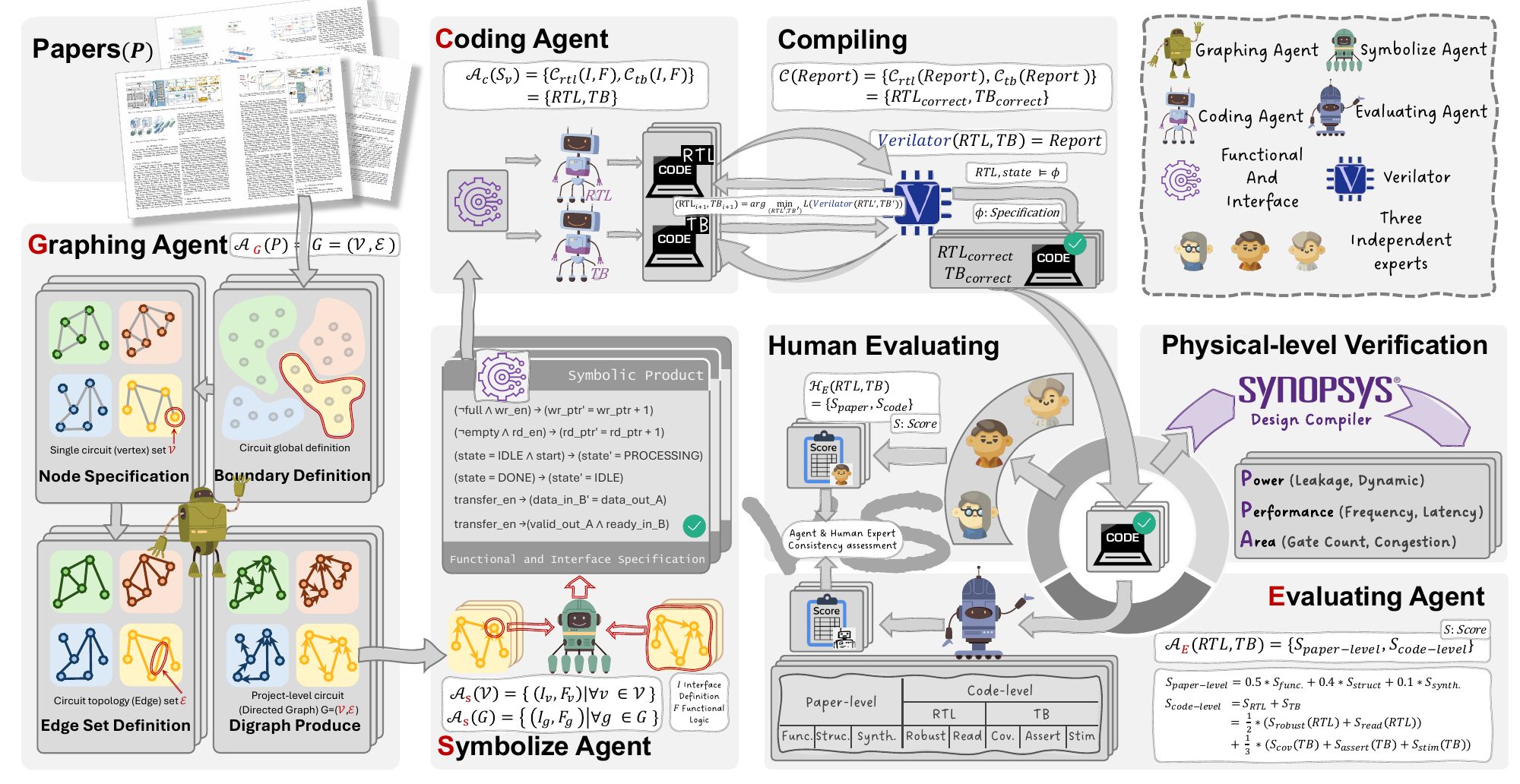}
    \vspace{-10pt}
    \caption{Overview of our ArchCraft framework with four phases. (1) Graphing Agent parses the paper into a formal architectural knowledge graph. (2) Symbolize Agent converts the graph nodes into a rigorous symbolic blueprint. (3)  Coding Agent decoupledly generates synthesizable RTL and TBs from the symbolic blueprint. (4) Automatic execution of compilation and checking of grammar and logical issues, collects error reports, holds errors and related codes accountable, and collaboratively corrects them within a continuous feedback loop. Finally, both LLM and human experts evaluate the generated code to assess the overall effectiveness of the system. The generated RTL code is then synthesized into a hardware implementation for physical-level evaluation.} 
    \label{fig:figure_2}
    \vspace{-10pt}
\end{figure*}

\section{ArchCraft}
In this section, we introduce ArchCraft, a Neural-Symbolic graph-based framework designed to overcome the challenges of hardware reproduction stemming from the scarcity of open-source materials. Unlike previous approaches, such as specialized large model fine-tuning or circuit-level agents, which depend on implicit neural representations and entail costly training processes and input prerequisites, our ArchCraft embeds a Mental Model. The generated circuit code supports comprehensive evaluation from functional to physical implementation(Sections \ref{subsec:Physical}). Prompts for the Agentic framework are in Appendix \ref{sec:prompts}.

\subsection{The Very Beginning}
\label{subsec:begining}
We utilize papers from the domain of architecture, particularly those related to ASICs, as input. In contrast to the industrial documents required in conventional workflows, academic papers represent a form of novel, rigorous, and high-quality documentation. Furthermore, compared to industrial documents used internally within engineering departments, academic papers are abundant and openly accessible to all in the relevant field for reading and study. However, a challenge with academic papers is that they can be more abstract, lacking descriptions of circuit-level design details. A more detailed comparison of these inputs is provided in the Appendix \ref{table: difference}. This Neural-Symbolic graph-based Agentic framework is therefore proposed to address this challenge. 

\subsection{Graphing Phase}
\label{subsec:graph}
The Graphing phase is orchestrated by the Graphing agent, $\mathcal{A}_{\textcolor{red}{\mathcal{G}}}$, taking an academic paper $P$ as input and produces a graph $\mathcal{G}$, 
where $G$ consists of a set of nodes $V$ and a set of edges $\mathcal{E}$:

\vspace{-2mm}
\begin{equation}
\mathcal{G} = (\mathcal{V}, \mathcal{E})
\end{equation}
\vspace{-5mm}

Where $\mathcal{V}$ stands for nodes, $\mathcal{E}$ stands for edges. This transformation is decomposed into a rigorous four-step sequence, where each step contributes specific components to the final graph structure.

\noindent\textbf{Scope and Theme}
First, the agent establishes the high-level context. This step defines the boundaries of the scope $S$ and the primary objectives or themes $T$ of the implementation, extracted directly from the paper's claims and contributions. Formally, this process can be described as:

\vspace{-2mm}
\begin{equation}
{\mathcal{A}_{\textcolor{red}{\mathcal{G}}}}_{1} : P \mapsto (S, T)
\end{equation}
\vspace{-5mm}

where ${\mathcal{A}_{\textcolor{red}{\mathcal{G}}}}_{1}$ denotes the agent responsible for scope establishment.


\noindent\textbf{Nodes and Static Links}
Based on the established scope $S$, the agent identifies the core components and their physical or hierarchical relationships. This step defines the set of nodes $\mathcal{V}$, representing hardware modules and the set of static, undirected edges $\mathcal{E}_{\text{static}}$, representing structural connections or module containment.
Formally, this process can be represented as a mapping:

\vspace{-2mm}
\begin{equation}
\mathcal{A}_{\mathcal{G},2} : (P, S) \mapsto (\mathcal{V}, \mathcal{E}_{\text{static}})
\end{equation}
\vspace{-5mm}

where ${\mathcal{A}_{\textcolor{red}{\mathcal{G}}}}_{2}$ denotes the agent responsible for nodes and static links construction.

\noindent\textbf{Direction and Internal Nodes}
With the nodes $\mathcal{V}$ defined, this step details the functional behavior and data pathways. The agent plans the graph's direction by defining directed edges $\mathcal{E}_{\text{dir}}$, representing dataflow and assigns a set of internal properties $\Phi$ (e.g., functionality, parameters) to each node $v \in \mathcal{V}$. Mathematically, this process is represented by:

\vspace{-2mm}
\begin{equation}
{\mathcal{A}_{\textcolor{red}{\mathcal{G}}}}_{3} (P, \mathcal{V}) \mapsto (\mathcal{E}_{\text{dir}}, \Phi)
\end{equation}
\vspace{-5mm}

\noindent in which ${\mathcal{A}_{\textcolor{red}{\mathcal{G}}}}_{3}$ is the agent for edge generation.

\noindent\textbf{Global Constraints}
Finally, the agent extracts system-wide requirements that apply to the entire graph. This includes global constraints $C$, such as clocking schemes, bus standards, or top-level performance targets, which are critical for ensuring architectural coherence. This extraction task is formalized as the following mapping:

\vspace{-2mm}
\begin{equation}
{\mathcal{A}_{\textcolor{red}{\mathcal{G}}}}_{4} (P) \mapsto C
\end{equation}
\vspace{-5mm}

\noindent Here, ${\mathcal{A}_{\textcolor{red}{\mathcal{G}}}}_{4}$ denotes the agent responsible for identifying and extracting the set of global constraints $C$ from the entire paper $P$.

The final graph is $\mathcal{G}=(\mathcal{V}, \mathcal{E})$. In this graph, the nodes $\mathcal{V}$ represent all registers and IO ports, while the edges $\mathcal{E}$ represent the dependency relationships between them. This composite graph, formed by the union of static edges $\mathcal{E}_{\text{static}}$ and directed edges $\mathcal{E}_{\text{dir}}$, is augmented by node attributes $\Phi$ and governed by global constraints $C$. This complete graph $\mathcal{G}$ serves as the formal architectural blueprint for all subsequent stages.

\subsection{Symboling Phase}
Following the construction of the architectural blueprint $\mathcal{G}=(\mathcal{V},\mathcal{E})$, the Symbolization phase, executed by the Symbolize Agent $\mathcal{A}_{\textcolor{red}{\mathcal{S}}}$, translates the internal logic and attributes of each graph node into a formal symbolic representation. As illustrated in the provided Figure \ref{fig:figure_2}, the agent's operation is defined at two levels:


\noindent\textbf{Graph-level Symbolization} From a global perspective, the agent's task is to define the overall specification for design, which is itself hierarchical. If the intent is to produce a top-level model specification for the entire graph $\mathcal{G}$, we write this mapping as:
\vspace{-1mm}
\begin{equation}
\mathcal{A}_{\textcolor{red}{\mathcal{S}}} : \mathcal{G} \mapsto (I_{\mathcal{G}},F_{\mathcal{G}}),
\qquad \text{where } \mathcal{G}=(\mathcal{V},\mathcal{E}).
\end{equation}
\vspace{-5mm}

\noindent This top-level specification $(I_{\mathcal{G}},F_{\mathcal{G}})$ is inherently defined by the composition of its internal parts.

Alternatively, the agent can be applied to a family of subgraphs (or subsystems) $\mathfrak{G} \subseteq \mathcal{P}(\mathcal{G})$, which represent the intermediate modules in the hierarchy. This is expressed as:

\vspace{-2mm}
\begin{equation}
\mathcal{A}_{\textcolor{red}{\mathcal{S}}}(\mathfrak{G}) = \{ (I_g, F_g) \mid g \in \mathfrak{G} \}.
\end{equation}
\vspace{-5mm}

\noindent This hierarchical decomposition continues until the process reaches the individual nodes, which require their own specific definitions.

\noindent\textbf{Node-level Symbolization}
For every node $v\in\mathcal{V}$, the agent $\mathcal{A}_{\textcolor{red}{\mathcal{S}}}$ produces an interface definition $I_v$ and a functional-logic specification $F_v$. This captures the complete symbolic product for each module.
The per-node mapping can be expressed as:

\vspace{-6mm}
\begin{equation}
\mathcal{A}_{\textcolor{red}{\mathcal{S}}} : \mathcal{V}\to\mathcal{I}\times\mathcal{F}, \qquad
v \mapsto (I_v,F_v),
\end{equation}
\vspace{-6mm}

\noindent and the image of the whole node set is

\vspace{-2mm}
\begin{equation}
\mathcal{A}_{\textcolor{red}{\mathcal{S}}}(\mathcal{V}) \;=\; \{(I_v,F_v)\mid v\in\mathcal{V}\}.
\end{equation}
\vspace{-5mm}

\noindent After the above two steps, we will get all levels of interfaces and functions of a project-level circuit design. This complete set of specifications, $\{(I, F)\}$, using the notation from the previous equations, this set is:

\vspace{-4mm}
\begin{small}
\begin{equation}
\{(I,F)\} = \mathcal{A}_{\mathcal{S}}(\mathcal{V}) \;\cup\; \mathcal{A}_{\mathcal{S}}(\mathfrak{G}) \;\cup\; \{(I_{\mathcal{G}}, F_{\mathcal{G}})\}
\end{equation}
\end{small}
\vspace{-5mm}

\vspace{-4mm}
\subsection{Coding and Compiling Phase}
This phase transforms the interface $I$ and functional specification $F$ into synthesizable RTL code and its corresponding TB.

\subsubsection{Coding}
The Coding Agent, $\mathcal{A}_{\textcolor{red}{\mathcal{C}}}$ takes the symbolic interface $I$ and functional logic $F$ for each module as its input and generates two distinct code artifacts: the RTL implementation $\mathcal{C}_{rtl}$ and the TB $\mathcal{C}_{tb}$.

\vspace{-4mm}
\begin{equation}
\begin{aligned}
\mathcal{A}_{\textcolor{red}{\mathcal{C}}}(I,F) &= (\mathcal{C}_{rtl}(I, F), \mathcal{C}_{tb}(I, F)) \\
                 &= (RTL, TB)
\end{aligned}
\end{equation} 
\vspace{-4mm}

RTL and TB are generated decoupledly, ensuring that the functional hardware logic and its verification environment are both directly grounded in the same formal specification.

\subsubsection{Compiling and Internal Feedback Loop}
The generated code artifacts \{$RTL, TB$\} are not assumed to be correct on the first attempt. They immediately enter a rigorous verification and refinement loop, as illustrated in the Figure \ref{fig:figure_2}.

\noindent\textbf{Verification:} The code is passed to an external simulator (e.g., \textcolor[HTML]{465B98}{Verilator}) for compilation and execution, which produces a simulation $Report$.

\vspace{-3mm}
\begin{equation}
\textcolor[HTML]{465B98}{Verilator}(RTL, TB) = Report
\end{equation} 
\vspace{-5mm}

\noindent\textbf{Internal Reflection:} The $Report$ is then fed back into a corrector function $\mathcal{C\_F}(Report)$. This function analyzes the report to identify discrepancies and determines if the current code is correct {\small $\{ RTL_{correct}, TB_{correct}\}$}.

\begin{small} 
\begin{equation}
\begin{aligned}
\mathcal{C\_F}(Report)
    &= (\mathcal{C}_{rtl}(Report), \mathcal{C}_{tb}(Report)) \\
    &\stackrel{?}{=} (RTL_{correct}, TB_{correct})
\end{aligned}
\end{equation} 
\end{small}
\vspace{-5mm}

\noindent\textbf{Iterative Refinement:} If errors are detected, e.g., $Report$ indicates failure, the framework activates an internal feedback loop. This iterative process, defined as $i+1$, seeks to find a corrected set of code ({\small $RTL', TB'$}) that minimizes the error loss function $L$ based on the output of the \textcolor[HTML]{465B98}{Verilator} as:

\begin{small} 
\begin{equation}
\begin{aligned}
\arg \min_{(RTL', TB')} 
                          L(\textcolor[HTML]{465B98}{Verilator}(RTL', TB'))
\end{aligned}
\end{equation}
\end{small}

This feedback cycle: Coding $\rightarrow$ Verilator $\rightarrow$ Report $\rightarrow$ Refinement repeats, continuously checking the RTL's state ($\Phi$) against the original specification. The loop ends only when the code passes all the checks, resulting in verified artifacts {\small $\{ RTL_{correct}, TB_{correct}\}$}.


\section{ArchSynthBench}
This section further introduces ArchSynthBench from three perspectives: the composition of the dataset (Section \ref{subsec:datasource}), the metrics and scoring methodologies for both machine and human expert evaluations on the benchmark (Section \ref{subsec:evaluation}),  and the physical implementation methods for the circuits (Section \ref{subsec:Physical}).

\subsection{Data Source}
\label{subsec:datasource}
ArchSynthBench consists of 50 distinct, closed-source hardware papers spanning 50 project-level circuits, and around \textbf{600} circuit blocks, which VerilogEval\cite{liu2023verilogeval} has only 156 circuit blocks. This represents a significant expansion in project-level complexity over module-centric benchmarks like VerilogEval. Beyond its scale, ArchSynthBench introduces comprehensive criteria for evaluating agent workflows. The full paper list, along with detailed paper's theme, is available in Appendix Table~\ref{tab:papers_by_year} and Figure \ref{fig:keyword_wordcloud}.


\subsection{Evaluating Phase}
\label{subsec:evaluation}
Following recent LLM-as-judge approaches \cite{gu2024survey, chen2024mllm}, to quantitatively assess the quality of the generated artifacts, we introduce a comprehensive, two-tiered evaluation system orchestrated by an Evaluating Agent $\mathcal{A}_{\textcolor{red}{\mathcal{E}}}$, specific scoring criteria are set to measure the codes, detailed in Appendix Section \ref{sec:scoring_criteria}. This phase takes the verified $RTL$ and $TB$ as inputs and computes two distinct scores $S$, in Figure \ref{fig:figure_2}: {\small $S_{paper-level}$} and {\small $S_{code-level}$}.


\noindent\textbf{Paper-level Score}
The $S_{\text{paper-level}}$ score measures the fidelity of the generated code to the original academic paper. It assesses how well the implementation reproduces the concepts described in the source text. This score is defined as a weighted sum of three sub-metrics: Functionality $S_{\text{Func.}}$, Structure $S_{\text{Struct.}}$, and Synthesizability $S_{\text{Synth.}}$, as follow:

\vspace{-4mm}
\begin{small}
\begin{equation}
S_{\text{paper-level}} = 0.5 \cdot S_{\text{func.}} + 0.4 \cdot S_{\text{struct.}} + 0.1 \cdot S_{\text{synth.}}
\end{equation}
\end{small}
\vspace{-6mm}

Detailed definitions for these sub-metrics are available in Appendix Table \ref{tab:rubric-rtl}.

\noindent\textbf{Code-level Score}
The $S_{\text{code-level}}$ score assesses the intrinsic quality and robustness of the generated code itself, independent of the source paper. It is the sum of the RTL Score $S_{RTL}$ and the TB Score $S_{TB}$.
The $S_{RTL}$ is the \textbf{average} of the design's Robustness $S_{\text{Robust.}}$ and Readability $S_{\text{Read.}}$.
The $S_{TB}$ is the \textbf{average} of the verification environment's Coverage $S_{\text{Cov.}}$, Assertions $S_{\text{Assert.}}$, and Stimulus $S_{\text{Stim.}}$.
Detailed evaluation rubrics for these five sub-metrics are provided in Appendix \ref{tab:rubric-rtl}, \ref{tab:rubric-tb}.

In addition, we have set up a penalty mechanism. If the generated code is not even in Verilog language, the code-level score will be further reduced by 10\%. These machine-generated scores are then used in conjunction with human expert evaluations to perform a consistency assessment, ensuring the automated metrics align with human judgment on design quality and correctness.

\subsection{Physical-level Verification}
\label{subsec:Physical}
Beyond functional correctness, a crucial aspect of our evaluation is assessing the physical-level feasibility and efficiency of the generated hardware. This step moves from simulation to synthesis, providing tangible metrics for the design's real-world viability.


As illustrated in \ref{fig:figure_2}, the functionally correct RTL code is fed into an industry-standard logic synthesis tool, which synthesizes the RTL into a gate-level netlist, from which we extract the essential PPA metrics:

\begin{itemize}[leftmargin=*]
\vspace{-2mm}
    \item \textbf{Power:} We analyze the estimated power consumption, including both static \textbf{Leakage} power and \textbf{Dynamic} power dissipated during operation.
    \vspace{-6mm}
    \item \textbf{Performance:} We evaluate the timing characteristics, primarily the operational \textbf{Latency}.
    \vspace{-2mm}
    \item \textbf{Area:} We measure the total design footprint, reported as the logical \textbf{Gate Count}, and assess the routing difficulty via the \textbf{Congestion} report.
\end{itemize}

This PPA analysis allows for a comparison against the results reported in the academic paper, providing the definitive measure of our framework's ability to reproduce not just the design's logic, but also its physical quality and efficiency.

\begin{table*}[htbp]
\centering
\resizebox{\textwidth}{!}{%
\setlength{\tabcolsep}{5.8pt}
\begin{tabular}{llcccccccc}
\toprule
\toprule
\multirow{2}{*}{Method} & \multirow{2}{*}{LLM} & \multicolumn{4}{c}{Paper-level} & \multicolumn{4}{c}{Code-level} \\
\cmidrule(lr){3-6} \cmidrule(lr){7-10} 
& & \multicolumn{1}{c}{Func.} & \multicolumn{1}{c}{Struc.} & \multicolumn{1}{c}{Synth.} & \multicolumn{1}{c}{\textbf{WA}} & \multicolumn{1}{c}{RTL} & \multicolumn{1}{c}{TB} & \multicolumn{1}{c}{\textbf{Comp.}} & \multicolumn{1}{c}{\textbf{WA}} \\
\midrule
Direct & Gemini 2.0 Flash & 15.00 & 28.00 & 27.50 & 21.45 & 36.83 & 25.11 & \ding{55} & 32.92 \\
Direct & Qwen3-Coder-480B & 27.50 & 28.13 & 28.75 & 27.88 & 46.46 & 43.82 & \ding{55} & 45.58 \\
Direct & GPT-4o  & 27.22 & 34.72 & 34.44 & 30.94 & 41.11.33 & 31.20 & \ding{55} & 42.54 \\
Direct & o3-mini & 31.00 & 33.50 & 35.50 & 32.45 & 57.33 & 43.31 & \ding{55} & 52.66 \\
\midrule
ChatDev & GPT-4o & 41.02 & 44.96 & 40.10 & 42.50 & 17.50 & 9.92 & \ding{55} & 14.97\\
VerilogCoder & o3-mini & 41.79 & 38.21 & 46.07 & 40.79 & 43.57 & 00.00 & \ding{55} & 34.21 \\
PaperCoder & o3-mini & 70.11 & 69.26 & 63.07 & 69.07 & 50.41 & 33.93 & \ding{55} & 44.91 \\
\midrule
ArchCraft & Gemini 2.0 Flash & 78.30 & 77.11 & 79.67 & 78.76 & 57.56 & 52.06 & \ding{51} & 55.72 \\
ArchCraft & Qwen3-Coder-480B & 77.28 & 80.28 & 70.13 & 78.57 & 66.71 & 68.40 & \ding{51} & 67.28 \\
ArchCraft & GPT-4o & 76.57 &  76.96 & 79.21 & 76.99 & 72.64 & 66.26 & \ding{51} & 70.51 \\
ArchCraft & o3-mini & \textbf{85.62} & \textbf{86.51} & \textbf{86.17} & \textbf{86.18} & \textbf{86.03} & \textbf{71.06} & \ding{51} & \textbf{81.04} \\
\midrule
\bottomrule
\end{tabular}
}
\vspace{-5pt}
\caption{Overall evaluation scores on ArchAynthBench, 50 project-level large circuits, over 500 circuit blocks, for Paper-level and Code-level metrics. ``Comp.'' denotes the Compilation Pass Status and ``WA'' denotes the Weighted Average score. The best performance is denoted in \textbf{bold}.}
\label{table:combined_evaluation_scores}
\vspace{-10pt}
\end{table*}

\section{Experiments}
In this section, we describe the experimental setup and results in detail.
\subsection{Experiment Settings}
\textbf{Baseline.} 
We propose the novel task of generating hardware RTL code from academic hardware papers. Given the absence of agentic workflow baselines specialized for this task, we compare our framework directly against LLMs. These LLMs include OpenAI o3-mini \cite{openai-o3-mini}, Gemini 2.0 Flash \cite{deepmind-gemini-models}, and Qwen3-Coder-480B-A35B-Instruct-FP8 \cite{qwen3technicalreport}, all of which generate RTL code in a single step from the raw paper input, devoid of any intermediate workflow or agentic collaboration, thus effectively acting as black-box generators.

For additional context and to evaluate the broader efficacy of our framework, we also benchmark its performance against existing work in the domain of automated software generation, notably ChatDev \cite{qian2023chatdev}, Paper2Code \cite{seo2504paper2code}, and VerilogCoder \cite{ho2025verilogcoder}. All aforementioned baselines, along with our proposed ArchCraft framework, are evaluated using our custom ArchSynthBench benchmark. Crucially, the RTL and TB code generated by all baselines are evaluated using the identical standards and the o3-mini evaluation agent applied to ArchCraft on ArchSynthBench, ensuring a fair and consistent comparison.

\noindent\textbf{Hardware Implementation Environment.} We conduct comprehensive hardware synthesis experiments using Synopsys DC, a widely adopted ASIC synthesis tool. All experiments are performed using Synopsys DC at 200MHz on SIMC 28nm technology on a high-performance server equipped with Intel Xeon Gold CPUs and 256 GB RAM under CentOS 7.

\subsection{Result}

\textbf{Main Result.}
Our experiment encompasses both paper- and code-level evaluation of all baselines, detailed code-level scores in all dimensions can be seen in the Appendix \ref{sec:code_level_detailed_scores}. The results in Table \ref{table:combined_evaluation_scores} demonstrate that ArchCraft outperforms all the baselines, including direct LLM and other agent frameworks such as ChatDev. These results highlight the superior capability of ArchCraft in generating project-level, synthesizable RTL code directly from research papers. Detailed examples of original machine-based evaluation scores can be found in the Appendix Section \ref{sec:original_scores}.

In both paper- and code-level evaluations, baselines using direct LLM generation consistently demonstrate poor performance. These methods only achieve around 30\% structural clarity and synthesisability scores. These severe limitations stem from their black-box, non-iterative generation approach. At the paper-level evaluation, they struggle with the comprehensive understanding and conceptual planning of circuit architectures described in the input papers, resulting in their failing to capture the intended design logic. At the code level, the absence of integration toolchains leads to weak iterative refinement and compilation feedback loops, with generated code failing to compile and even failing logic synthesis.

Similarly, prior multi-agent software frameworks, such as ChatDev, PaperCode, and VerilogCoder, are also incapable of generating synthesizable RTL code. While these agent frameworks have been demonstrated to outperform black-box LLMs at the paper or code level, their inherent design presents limitations. They either lack the capability for project-level RTL code generation or impose stringent input requirements, rendering them ineffective at extracting circuit architectures from abstract paper inputs. Consequently, they underperform our ArchCraft in the paper-to-project-level task. Moreover, the design paradigms of ChatDev and PaperCode inherently lack crucial integration with EDA toolchains, which entirely preclude the compilation and iterative modification of the generated hardware descriptions.

In contrast, ArchCraft stably generates compilable RTL and TB code across all LLM backbones and achieves significantly higher scores across all evaluation metrics. This superior performance is attributed to ArchCraft's Neural-Symbol Graph-based framework, which is capable of sketching circuit structures from paper documents, combined with its compiler toolchain-integrated code traceability and error-correction feedback mechanisms. By analyzing fundamental circuit elements, integrating key error analysis, and implementing targeted corrective measures throughout the process, ArchCraft effectively bridges the gap between closed-source, high-level paper specifications and functionally verifiable, deployable hardware implementations. Due to space limitations, ArchCraft's rectifying analysis is shown in Appendix \ref{sec:rectifying_analysis}.

\begin{table}[t]
\centering
\resizebox{\columnwidth}{!}{
    \begin{tabular}{llcc}
    \toprule
    Method & LLM & Paper-level & Code-level \\
    \midrule
    Direct & Gemini 2.0 Flash & 22.17 & 37.03  \\
    Direct & Qwen3-Coder-480B & 22.00 & 44.44 \\
    Direct & GPT-4o & 26.54 & 47.73 \\
    Direct & o3-mini & 29.67 & 50.74  \\
    \midrule
    ChatDev & GPT-4o & 26.44 & 10.89 \\
    VerilogCoder & o3-mini & 31.27 & 48.89  \\
    PaperCoder & o3-mini & 49.68 & 47.33  \\
    \midrule
    ArchCraft & Gemini 2.0 Flash & 75.24 & 70.11 \\
    ArchCraft & Qwen3-Coder-480B & 76.90 & 78.54 \\
    ArchCraft & GPT-4o & 75.59 & 71.47 \\
    ArchCraft & o3-mini & 79.13 & 84.33 \\
    \bottomrule
    \end{tabular}
} 
\vspace{-5pt}
\caption{Human expert evaluation scores from both paper and code levels.}
\vspace{-10pt}
\label{table:ablation}
\end{table}


\textbf{Human Expert Assessments.}
To rigorously validate the accuracy of our automated evaluation metrics, we contract three independent human engineering experts to assess the code quality generated by the direct method, other frameworks, and our ArchCraft. The evaluation criteria and scoring methodology used by these human experts have the same dimension as those used in the machine-based evaluation assessment. As presented in Table \ref{table:ablation}, the results unequivocally demonstrate that ArchCraft achieves the best ranking among all compared methodologies. For detailed original human evaluation scores and the Pearson correlation coefficient between the scores assigned by human experts and those derived from our machine-based evaluations, see the Appendix \ref{sec:Pearson_correlation} and \ref{sec:original_scores}.

\textbf{Hardware Synthesis Evaluation.}
We conducted synthesis experiments on the RTL circuits for 60 modules, randomly sampled from ten papers in ArchSynthBench, using Synopsys's DC software. Due to space constraints, the PPA report for a subset of these circuits is presented in Appendix \ref{sec:physical_implementation}. Table \ref{tab:synthesis_success_rate} summarizes the synthesis success rate, where success is defined as the proportion of successfully synthesized designs relative to all tested designs, excluding top-level designs and those with known syntactical errors. The results indicate that the majority of designs generated by our framework are synthesizable, meet timing constraints, and are free of violations, which confirms their structural integrity and practical feasibility for downstream ASIC implementation. The few observed synthesis failures were primarily attributed to syntactical issues or port definition mismatches, highlighting a clear direction for improving generation robustness in future work.

\begin{table}[t]
\centering
\resizebox{\columnwidth}{!}{%
\footnotesize
\setlength{\tabcolsep}{4pt}
\begin{tabular}{lclc}
\toprule
Paper Title &  (Success/Total) & Paper Title &  (Success/Total) \\
\midrule
FlightVGM   & 6 / 8  & PBN           & 5 / 5 \\
Flex-EGAI   & 4 / 5  & SPARK         & 7 / 7 \\
LC-MAC      & 3 / 4  & ST-Purning    & 5 / 5 \\
INSPIRE     & 4 / 6  & MASL\_AFU     & 5 / 7 \\
bitWAVE     & 5 / 6  & ViTA          & 7 / 8 \\

\bottomrule
\end{tabular}
}
\vspace{-5pt}
\caption{Synthesis success rates across part papers evaluated.}
\vspace{-10pt}
\label{tab:synthesis_success_rate}
\end{table}

\begin{table}[t]
\centering
\resizebox{\columnwidth}{!}{%
\footnotesize
\setlength{\tabcolsep}{4pt}
\begin{tabular}{lccc}
\toprule
Design & Power($\mu$W) & Slack/CPD(ns) & Area($\mu m^2$) \\
\midrule
control\_unit         & 170.1   & 3.92/0.93 & 954.68   \\
im2col\_pack\_engine  & 1039.9  & 2.31/2.52 & 5758     \\
interface             & 603.3   & 3.44/1.41 & 3945.11  \\
memory\_controller    & 939.7   & 2.85/2.06 & 6200.41  \\
pe\_array             & 5262.2    & 1.04/3.81 & 32345.91 \\
spark\_decoder        & 109.0   & 3.88/0.98 & 568.81   \\
spark\_encoder        & 324.4   & 3.47/1.38 & 1676.51  \\
\bottomrule
\end{tabular}
}
\vspace{-5pt}
\caption{Synthesis PPA results for SPARK designs using Synopsys DC. All the designs have no time violations. CPD is the abbreviation of Critical Path Delay. }
\label{tab:spark_synthesis}
\vspace{-10pt}
\end{table}

As a case study, Table \ref{tab:spark_synthesis} presents the post-synthesis PPA results made by ArchCraft with o3-mini as backbone, for designs derived from the SPARK \cite{liu2024spark}. All the implementations of the SPARK paper successfully pass synthesis with positive slack, confirming their feasibility for timing closure at the target frequency. Notably, the logic synthesizes a single PE that occupies approximately 4k \textmu $\text{m}^2$, with a power consumption of 0.74 mW. While an area discrepancy exists due to specific design details compared to the original SPARK paper, our generated circuit design reproduces the relative scale and power characteristics expected for compute-intensive accelerators, considering the overall relationships between the various designs. Furthermore, the control and interface logic demonstrates sub-mW power consumption and compact area footprints, consistent with their functional roles within the SPARK architecture. These empirical results underscore the efficacy of our ArchCraft framework in faithfully translating high-level architectural descriptions into synthesizable RTL implementations with practical PPA characteristics.


\section{Conclusion}

This paper introduces ArchCraft, a framework that captures the architectural blueprint using formal graphs and defines functional specifications with symbols, transforming abstract architectural descriptions from academic papers into synthesizable Verilog projects with functional verification. In addition, we constructed ArchSynthBench, a benchmark specifically designed for synthesizing project-level circuits from closed-source academic papers describing especially for artificial intelligence accelerators. Extensive experimental results demonstrate that ArchCraft achieves state-of-the-art performance in machine evaluations, human expert assessments, and physical-level verification, while exhibiting robustness and practicality across different LLM backbones.We expect that the proposed ArchCraft will contribute to accelerating scientific progress by lowering the barriers to hardware replication and enhancing research reproducibility and innovation.

\bibliography{custom}

\clearpage
\appendix

\section{Ethics Statement}
The primary objective of this work is to automate the replication of experiments detailed in existing, publicly available research papers. While the methodologies themselves are drawn from the public domain, which generally implies transparency, it is important to acknowledge the potential for data leakage associated with the use of our system. Users should therefore be mindful of this possibility, particularly when the replication process might involve sensitive datasets or generate intermediate results that could inadvertently disclose information.

\section{Prompts}
\label{sec:prompts}
In this section, we provide a detailed exposition of the four phases within our framework. The specific prompts utilized at each stage of this process are presented in Section \ref{sec:graphing_prompt} is Graphing Prompts, Section \ref{sec:symbolize_prompt} is Symbolize Prompts, Section \ref{sec:coding} is Coding Prompts, including original coding prompts and rectifying. 

\renewcommand{\dblfloatpagefraction}{.5}
\begin{figure}[htbp]
\begin{tcolorbox}[
    colback=white,
    colframe=black,
    title=\textbf{Prompts for graphing papers. I}
]
\textbf{System Role Prompt:}\\
You are an expert hardware engineer and RTL designer with a deep understanding of digital design and hardware implementation.\\
You will receive a research paper in <paper\_format> format.\\
Your task is to create a detailed and efficient plan to implement the hardware described in the paper.\\
This plan should align precisely with the paper's hardware architecture, timing requirements, and performance metrics.\\
The plan must be clear, structured, and focused on hardware implementation details.

\textbf{Task Prompt:}\\
1. We want to implement the hardware described in the attached paper.\\
2. The authors did not release any official RTL code, so we have to plan our own implementation.\\
3. Before writing any Verilog code, please outline a comprehensive plan that covers:\\
   - Key details from the paper's Hardware Architecture.\\
   - Important aspects of Implementation, including module hierarchy, interfaces, timing requirements, and verification strategy.\\
4. The plan should be as detailed and informative as possible to help us write the final RTL code later.

\textbf{Instruction Prompt:}\\
The response should give us a strong roadmap for hardware implementation, making it easier to write the RTL code later.
\end{tcolorbox}

\caption{Prompts for graphing papers.}
\label{fig:graphing_prompt_1}
\end{figure}

\subsection{Graphing Phase}
\label{sec:graphing_prompt}
We show 4 steps' prompts in our Graphing Phase. Figure \ref{fig:graphing_prompt_1} is Define Graph Scope and Theme. Figure \ref{fig:graphing_prompt_2} is Define Graph Nodes and Static Links. Figure \ref{fig:graphing_prompt_3} is Plan Graph Direction and Internal Node Properties. Figure \ref{fig:graphing_prompt_4} is Add Global Constraints.

\subsection{Symbolize Phase}
\label{sec:symbolize_prompt}
We show 3 steps' prompts in our Graphing Phase. Figure \ref{fig:symbolize_prompt} is Interface Protocol Design, Internal Logic Design,and Verification \& Implementation Plan.

\subsection{Coding Phase}
\label{sec:coding}
In this section, we demonstrated the decoupling generation of RTL and TB code in Figure \ref{fig:coding_prompt}, and the prompt used in the rectifying process of the generated code in Figure \ref{fig:rectifying_prompt}.


\renewcommand{\dblfloatpagefraction}{.7}
\begin{figure}[htbp]
\begin{tcolorbox}[
    colback=white,
    colframe=black,
    title=\textbf{Prompts for graphing papers. II}
]
\textbf{General Prompt:} \\
Your goal is to create a concise, usable, and complete hardware system design for implementing the paper's method. Use appropriate hardware design practices and keep the overall architecture modular. \\
Based on the plan for implementing the paper's main method, please design a concise, usable, and complete hardware system. Keep the architecture modular and make effective use of standard hardware design patterns.

\textbf{Format Prompt:} \\
Please note that this is just an example and in the processing of the actual module, there is no need to strictly adhere to the file names and the number of modules therein. Please refer to the actual analysis results.

\textbf{Action Prompt:} \\
Follow the instructions of the node to generate the output and ensure that it adheres to the format example. Please remember that the file list and module names should be adjusted according to the actual designed functions and do not have to strictly follow the examples.
\end{tcolorbox}

\caption{Prompts for graphing papers.}
\label{fig:graphing_prompt_2}
\end{figure}

\renewcommand{\dblfloatpagefraction}{.7}
\begin{figure}[htbp]
\begin{tcolorbox}[
    colback=white,
    colframe=black,
    title=\textbf{Prompts for graphing papers. III}
]
\textbf{General Prompt:} \\
Your goal is to break down tasks according to PRD/technical design, generate a task list, and analyze task dependencies. \\
You will break down tasks, analyze dependencies.\\
You outline a clear PRD/technical design for implementing the paper's hardware method.\\
Now, let's break down tasks according to PRD/technical design, generate a task list, and analyze task dependencies.\\
The Logic Analysis should not only consider the dependencies between modules but also provide detailed descriptions to assist in writing the RTL code needed to implement the paper.

\textbf{Format Prompt:} \\
Please note that this is just an example and in the processing of the actual module, there is no need to strictly adhere to the file names and the number of modules therein. Please refer to the actual analysis results

\textbf{Action Prompt:} \\
Follow the node instructions above, generate your output accordingly, and ensure it follows the given format example.
\end{tcolorbox}

\caption{Prompts for graphing papers.}
\label{fig:graphing_prompt_3}
\end{figure}

\renewcommand{\dblfloatpagefraction}{.7}
\begin{figure}[htbp]
\begin{tcolorbox}[
    colback=white,
    colframe=black,
    title=\textbf{Prompts for graphing papers. IV}
]
\textbf{General Prompt:} \\
You write elegant, modular, and maintainable RTL code. Adhere to hardware design guidelines.
Based on the paper, plan, and design specified previously, follow the "Format Example" and generate the code. 
Extract the hardware details from the above paper (e.g., clock frequency, data width, memory size, etc.), follow the "Format example" and generate the code. 
DO NOT FABRICATE DETAILS — only use what the paper provides.
You must write `config.yaml`.
\end{tcolorbox}

\caption{Prompts for graphing papers.}
\label{fig:graphing_prompt_4}
\end{figure}

\begin{figure*}[htbp]
\begin{tcolorbox}[
    colback=white,
    colframe=black,
    title=\textbf{Prompts for symbolize. I}
]
\textbf{System Role Prompt:}\ You are an expert hardware interface designer. Your task is to take a high-level module definition and design its detailed physical interfaces and protocols.

\textbf{Task Prompt:} Based on the high-level plan and the module list, refine the interfaces for the module: [Module\_Name\_From\_Graphing].

You must define:\\
Handshake Protocols: Specify the exact valid/ready or ack/req logic for each data port.\\ Memory Interfaces: If this module connects to memory, define the protocol (e.g., AXI-Lite, AXI-Stream, or simple BRAM interface) including all necessary signals (ar\_addr, ar\_valid, aw\_addr, w\_data, b\_ready, etc.). \\ 
Detailed Timing: Specify any multi-cycle signal behaviors or interface timing constraints not covered by the global config. \\
Control Signals: Define any new control signals required for this specific module's operation (e.g., start\_processing, op\_mode, done\_tick).\\

\end{tcolorbox}

\begin{tcolorbox}[
    colback=white,
    colframe=black,
    title=\textbf{Prompts for symbolize. II}
]
\textbf{System Role Prompt:}\ You are an expert RTL micro-architect. Your task is to design the internal control and data path logic for a given hardware module based on its function and interfaces.

\textbf{Task Prompt:} Design the internal micro-architecture for module:[Module\_Name\_From\_Graphing].

You must provide two separate but related designs:

\textbf{State Machine (FSM) Design:} \\
Identify FSMs: List all required Finite State Machines. \\
Define States: For each FSM, list its states.\\
Define Transitions: Describe the logic that causes transitions between states.\\
Control Logic: Specify the control signals generated in each state. \\
Reset Behavior: Define the FSM's state upon rst\_n assertion. 

\textbf{Data Path Design:} \\
Data Flow: Describe the path data takes through the module, from input ports to output ports. \\
Key Components: List the necessary data path components.\\
Pipeline Stages: If pipelining is required, define the logic for each stage. \\
Data Formats: Specify any changes in data format. \\
\end{tcolorbox}

\begin{tcolorbox}[
    colback=white,
    colframe=black,
    title=\textbf{Prompts for symbolize. III}
]
\textbf{System Role Prompt:} You are an expert verification engineer and physical designer. Your task is to create a test plan and define implementation constraints for a given hardware module.

\textbf{Task Prompt:} Create the verification plan and define implementation constraints for module: [Module\_Name\_From\_Graphing].

\textbf{Verification Requirements:} \\
Test Scenarios: Define key scenarios to test. \\
Coverage Points: List critical coverage points. \\
Test Vectors: Provide 2-3 example test vectors. 

\textbf{Implementation Constraints:} \\
Module-Specific Constraints: Define constraints specific to this module that override or add to the global config .\\
Optimization Goals: Specify the priority for this block.
\end{tcolorbox}
\caption{Prompts for symbolize.}
\label{fig:symbolize_prompt}
\end{figure*}

\clearpage
\begin{figure*}[htbp]
\begin{tcolorbox}[
    colback=white,
    colframe=black,
    title=\textbf{Prompts for coding. I}
]
\textbf{System Role Prompt:}\\
You are an expert researcher and software engineer with a deep understanding of experimental design and reproducibility in scientific research.\\
You will receive a research paper in format, an overview of the plan, a Design in JSON format consisting of "Implementation approach", "File list", "Data structures and interfaces", and "Program call flow", followed by a Task in JSON format that includes "Required packages", "Required other language third-party packages", "Logic Analysis", and "Task list", along with a configuration file named "config.yaml". \\
Your task is to write code to reproduce the experiments and methodologies described in the paper. \\
The code you write must be elegant, modular, and maintainable, adhering to Google-style guidelines. \\
The code must strictly align with the paper's methodology, experimental setup, and evaluation metrics.\\ 
Write code with triple quotes.

\textbf{RTL Coding Prompt:}\\
Based on the planning, design, and tasks of the thesis and the configuration file "config.yaml" specified earlier, write the code in accordance with the "Format Example".

We have done the file list.
Next, you must only write the todo file name.\\
1. Only one file: Do your best to achieve this with only one file.\\
2. Complete code: Your code will be part of the entire project, so please implement complete, reliable, and reusable code snippets. \\
3. Follow the design: You must adhere to "Data Structures and Interfaces". Don't change any design. Do not use public member functions that do not exist in the design.\\
4. Carefully check that no necessary classes/functions are missing in this file.\\
5. Before using an external variable/module, please make sure to import it first.\\
6. Write down every detail of the code and do not leave any tasks to be done.\\
7. Reference configuration: You must use the configuration in "config.yaml". Do not forge any configuration values.
\\
\end{tcolorbox}

\begin{tcolorbox}[
    colback=white,
    colframe=black,
    title=\textbf{Prompts for coding. II}
]
\textbf{TB Coding Prompt:}\\
You are an expert who can automate the RTL by using a fully autonomous toolchain. You are invited to teach us how to create chips by using a fully autonomous toolchain for digital layout generation across die sizes, process nodes, and foundry options.\\
Your specific task now is to write a complete and executable TB for the Verilog module named $`${module\_name}$`$. You must leverage the provided detailed module analysis to infer the module's interface and write a comprehensive and accurate TB.
 \\
Here is the detailed analysis of the module and its context from our autonomous toolchain: ${module\_analysis\_content}$\\
In addition, here are a few things you should notice: ${rules}$
\end{tcolorbox}
\caption{Prompts for coding.}
\label{fig:coding_prompt}
\end{figure*}

\clearpage
\begin{figure*}[htbp]
\begin{tcolorbox}[
    colback=white,
    colframe=black,
    title=\textbf{Prompts for Rectifying. I}
]
\textbf{Error Finding System Role Prompt:}\\
You are an experienced digital circuit engineer, proficient in Verilog/SystemVerilog design, simulation and debugging. \\
Your task is to analyze the provided Verilog simulation logs, identify the root cause of errors or warnings, and determine which file (such as' rtlfile.v 'or' testbenchfile.v ') is the main responsible for the problem. \\
Summarize the core lessons learned from this mistake in a concise sentence, with the focus on proposing a universal coding guideline to prevent similar mistakes from happening in the future.
\end{tcolorbox}

\begin{tcolorbox}[
    colback=white,
    colframe=black,
    title=\textbf{Prompts for Rectifying. III}
]
\textbf{Error Analysis System Role Prompt:}\\
You are an experienced digital circuit engineer, proficient in Verilog/SystemVerilog design, simulation and debugging.\\
Your task is to conduct an in-depth analysis of the provided Verilog simulation logs, RTL codes, and TB codes, identify the root causes of errors or warnings, and generate a detailed, fixe-oriented internal analysis result.\\
This analysis should clearly point out where the problem lies, why it occurs, and the potential methods and considerations for solving it (for example, if it involves width, it is necessary to consider how to correctly crop or expand, or whether the variable definition needs to be modified).\\
Please output your analysis results in JSON format, including the internal analysis text field.
\end{tcolorbox}

\begin{tcolorbox}[
    colback=white,
    colframe=black,
    title=\textbf{Prompts for Rectifying. III}
]
\textbf{Error Fixing System Role Prompt:}\\
You are an experienced digital circuit engineer, proficient in Verilog/SystemVerilog design, simulation and debugging.\\
Your task is to fix the Verilog code based on the provided original RTL/TBcode, simulation logs, and detailed internal analysis results .\\
Only modify the necessary code in the problematic files (RTL code and/or TB code) to solve the problem. If a file does not need to be modified, please return its original content.
The code should be completely and correctly formatted in the verilog code block.\\
Please output your reply in JSON format.
\end{tcolorbox}

\caption{Prompts for rectifying.}
\label{fig:rectifying_prompt}
\end{figure*}

\clearpage

\section{Differences between Paper and Document}
In this subsection, we provide a detailed comparison of the differences \ref{table: difference} between academic papers and industrial design documents as inputs to the ArchCraft framework. This distinction represents the foundational starting point for the entire ArchCraft process and the design motivation that led to the creation of its Neural-Symbolic Graph-based Framework.

\begin{table*}[htbp] 
\centering
\label{tab:feature-comparison}
\begin{tabularx}{\textwidth}{@{} X l c c @{}}
\toprule
\multicolumn{2}{l}{\textbf{Feature / Attribute}} & \textbf{Academic Paper} & \textbf{Industrial Doc.} \\
\midrule
\textbf{Accessibility} & Publicly Searchable   & \ding{51} & \ding{55} \\
                       & Abundant in Quantity    & \ding{51} & \ding{55} \\
                       & Internally Confidential & \ding{55} & \ding{51} \\
\midrule
\textbf{Content Focus} & Core Algorithm / Theory & \ding{51} & (\ding{51}) \\ 
& Rigorous Validation / Proofs & \ding{51} & \ding{55} \\
& Detailed Circuit-level Specs & \ding{55} & \ding{51} \\
& Implementation Details (APIs, etc.) & \ding{55} & \ding{51} \\
& Edge Cases              & \ding{55} & \ding{51} \\
\midrule
\textbf{Form}          & Standardized Structure  & \ding{51} & \ding{55} \\
& High-Level Abstraction  & \ding{51} & \ding{55} \\
& Low-Level Granularity   & \ding{55} & \ding{51} \\
\midrule
\textbf{Lifecycle}     & Static / Archived       & \ding{51} & \ding{55} \\
             & Dynamic / Living Document & \ding{55} & \ding{51} \\
    
\bottomrule
\end{tabularx}
\caption{Feature-based Comparison of Academic Papers and Industrial Design Documents.}
\label{table: difference}
\end{table*}

\section{Scoring Criteria }
\label{sec:scoring_criteria}
To support the quantitative and qualitative analyses presented in the main text, this appendix provides the detailed evaluation rubrics used to assess the quality of the generated hardware designs. These rubrics are explicitly divided into two parts: criteria for the design implementation (RTL) and criteria for its verification environment (Testbench).

\textbf{Table~\ref{tab:rubric-rtl}} details the evaluation rubric for the \textbf{RTL Design}. This rubric is structured around five core dimensions:

\begin{itemize}[leftmargin=*]
\vspace{-2mm}
    \item \textbf{Functional Correctness:} Assesses whether the design accurately implements the required algorithms and logic.
    \vspace{-2mm}
    \item \textbf{Robustness:} Examines the design's handling of boundary conditions, exceptions, and timing issues (e.g., CDC).
    \vspace{-2mm}
    \item \textbf{Structural Fidelity:} Evaluates the code's modularity, structural clarity, and naming conventions.
    \vspace{-5mm}
    \item \textbf{Synthesis Compatibility:} Ensures the code is synthesizable and evaluates the reasonableness of its resource utilization.
    \vspace{-2mm}
    \item \textbf{Readability:} Assesses the quality of comments and documentation, which relates to design maintainability.
\end{itemize}

Correspondingly, \textbf{Table~\ref{tab:rubric-tb}} presents the evaluation rubric for the \textbf{TB Design}. This rubric focuses on four key aspects:
\begin{itemize}[leftmargin=*]
\vspace{-2mm}
    \item \textbf{Test Case Coverage:} Evaluates whether the stimulus adequately covers functional paths, boundaries, and corner cases.
    \vspace{-2mm}
    \item \textbf{Robustness:} Focuses on whether the assertions are accurate, effective, and stable under various stimuli.
    \vspace{-2mm}
    \item \textbf{Stimulus Generation Quality:} Assesses the use of directed, random, and constrained-random strategies.
    \vspace{-2mm}
    \item \textbf{Maintainability:} Examines the clarity, extensibility, and debuggability of the testbench code itself.
\end{itemize}

As mentioned in the main text, each dimension is scored out of 100 points. The specific questions listed within these tables provide the basis for fine-grained scoring and qualitative feedback by human evaluators.

\begin{table*}[htbp]
\centering 
    \begin{tabular*}{\textwidth}{@{} l @{\extracolsep{\fill}} >{\raggedright\arraybackslash}p{0.7\textwidth} @{}}
    \toprule
    \addlinespace 
    \textbf{Functional Correctness} 
        & \textbullet~Does the RTL correctly implement the key algorithms, logic, or functional points described in the logical analysis? Is the functionality complete? \par \medskip 
          \textbullet~Does it faithfully reproduce the architecture, dataflow, and bit-width details specified in the logical analysis? \\
    \addlinespace
    
    \textbf{Robustness}
        & \textbullet~Does the RTL handle all input cases, boundary conditions, and potential exceptions (e.g., divide-by-zero, overflow, illegal inputs)? \par \medskip
          \textbullet~Is the reset logic clear and effective? Is Clock Domain Crossing (CDC) handled correctly (if applicable)? \par \medskip
          \textbullet~Is the design complete, unambiguous, and synthesizable without errors? \\
    \addlinespace
    
    \textbf{Structural Fidelity}
        & \textbullet~Is the code structure clear? Does it adhere to the logical layering from the analysis (if mentioned)? \par \medskip
          \textbullet~Is the module partitioning reasonable and the naming convention consistent? \\
    \addlinespace
    
    \textbf{Synthesis Compatibility}
        & \textbullet~Can the design be successfully synthesized by mainstream tools? Does it avoid non-synthesizable constructs? \par \medskip
          \textbullet~Is the resource utilization reasonable and aligned with the goals set in the logical analysis? \\
    \addlinespace
    
    \textbf{Readability}
        & \textbullet~Does the code have sufficient comments? Are signal names clear and meaningful? \par \medskip
          \textbullet~Does it include auxiliary documentation (e.g., parameter descriptions, structural comments)? \\
    
    \bottomrule
    \end{tabular*}
    
    \caption{Evaluation Rubric for RTL Design (100 points per dimension)}
    \label{tab:rubric-rtl}
\end{table*}

\begin{table*}[!h]
\centering
    \begin{tabular*}{\textwidth}{@{} l @{\extracolsep{\fill}} >{\raggedright\arraybackslash}p{0.7\textwidth} @{}}
    \toprule
    
    \addlinespace
    \textbf{Test Case Coverage}
        & \textbullet~Does the TB generate sufficient and representative test cases to cover all functional paths of the RTL? \par \medskip
          \textbullet~Does it cover boundary conditions, typical values, random values, and corner cases of the input space? \par \medskip
          \textbullet~Does it consider various timing combinations of control signals (e.g., reset, enable)? \\
    \addlinespace
    
    \textbf{Robustness}
        & \textbullet~Do the assertions (`assert`) accurately reflect the RTL's expected behavior? \par \medskip
          \textbullet~Are assertions granular enough to effectively capture errors in outputs or internal states? \par \medskip
          \textbullet~Are assertions robust, remaining stable and correct under varying or randomized stimulus? \par \medskip
          \textbullet~Are there checks on key intermediate signals, not just the final outputs? \\
    \addlinespace
    
    \textbf{Stimulus Generation Quality}
        & \textbullet~Is the method for generating input stimulus logical and flexible? \par \medskip
          \textbullet~Does it use a mix of directed, random, and constrained-random stimulus to enhance coverage? \par \medskip
          \textbullet~Is the generation of clock and reset signals compliant with the design specificaåtion? \\
    \addlinespace
    
    \textbf{Maintainability}
        & \textbullet~Is the testbench code clear, well-structured, and easy to understand? \par \medskip
          \textbullet~Is it easy to modify, extend with new test cases, or debug? \par \medskip
          \textbullet~Is the code free of redundancy and unnecessary complexity? \\
    
    \bottomrule
    \end{tabular*}
    \caption{Evaluation Rubric for TB Design (100 points per dimension)}
    \label{tab:rubric-tb}
\end{table*}

\section{Paper in ArchSynthBench Details}
\label{sec:appendix_paper_details}
In this subsection, we detail the corpus of academic papers that constitutes the \textbf{ArchSynthBench} benchmark. These papers serve as the high-level, abstract design specifications that our framework, ArchCraft, aims to interpret and implement. The complete list of these papers is presented in \textbf{Table~\ref{tab:papers_by_year}}, which is organized to highlight the breadth and timeliness of our curated corpus.

The table provides three key pieces of information for each entry:

\begin{itemize}[leftmargin=*]
\vspace{-2mm}
    \item \textbf{Year:} The publication year of the paper. This chronological organization (spanning 2022 to 2025) demonstrates that the benchmark is contemporary and built upon state-of-the-art research in hardware acceleration.
    \vspace{-2mm}
    \item \textbf{Paper:} The common acronym or a short name for the paper, along with its bibliographic citation key (e.g., \texttt{\cite{zhao2022fpga}}). This provides full transparency and traceability, allowing researchers to reference the original design specifications used in our evaluation.
    \vspace{-2mm}
    \item \textbf{Theme} A set of descriptive terms summarizing the paper's core technical contribution, target application, or key architectural paradigm.
\end{itemize}

An analysis of the \textbf{Theme} column reveals the extensive diversity of our benchmark. The corpus was intentionally curated to move beyond simple, canonical designs and to reflect the complexity of the modern hardware landscape. The topics span:

\begin{itemize}[leftmargin=*]
\vspace{-2mm}
    \item \textbf{Classic Acceleration Domains:} Such as CNN Accelerators (\texttt{CNN Accelerator}), Approximate Computing (\texttt{Approximate Computing / BNN}), and efficient inference (\texttt{Efficient NN Inference}).
    \vspace{-2mm}
    \item \textbf{Modern Neural Architectures:} Including dedicated hardware for Transformers (\texttt{FPGA Transformer}), Vision Transformers (\texttt{Vision Transformer / Accelerator}), and Large Language Models (\texttt{LLM Inference / PIM / CXL}).
    \vspace{-2mm}
    \item \textbf{Specific Hardware Techniques:} Such as \texttt{LUT-based NN}, \texttt{Sparsity}, \texttt{Quantization}, and Accelerator-in-Memory (\texttt{AiM}).
    \vspace{-2mm}
    \item \textbf{Emerging Paradigms:} The benchmark even includes forward-looking topics like CXL-based memory systems (\texttt{Memory Systems / CXL}) and Quantum Computing control (\texttt{Quantum Computing / Control}).
\end{itemize}

In summary, the paper corpus detailed in Table~\ref{tab:papers_by_year} provides a rich, challenging, and representative set of tasks. This diversity is essential for robustly evaluating the capability of any high-level synthesis agent to generalize across different domains and effectively bridge the gap from abstract textual specifications to concrete hardware implementations.

\begin{table*}[htbp]
\centering
\begin{tabular}{@{}cll@{}}
\toprule
\midrule
\textbf{Year} & \multicolumn{2}{l}{\textbf{Paper Name}} \\
\midrule
\multirow{5}{*}{2022} & OBS-TA \cite{zhao2022fpga} & Efficient Compression \cite{yan2022area} \\
 & LC-MAC \cite{li2022low} & Comprehensive Evaluation \cite{juracy2022comprehensive} \\
 & ST-Purning \cite{huang2022structured} & Approx. Arch. \cite{liu2022reconfigurable} \\
 & COMPAQT \cite{maurya2022compaqt} & NVP \cite{liu2022nvp} \\
 & Cryo-CMOS Transmon \cite{tien2022cryo} & GDDR6-Based AiM \cite{kwon20221ynm} \\
\addlinespace
\midrule
\multirow{3}{*}{2023} & AFIES \cite{ji2023fully} & Accurate Binary-Stochastic \cite{zhang2023high} \\
 & Efficient Multipliers \cite{sayadi2023two} & High-Precision Softmax \cite{zhang2023high} \\
 & FP Tensor Core \cite{venkataramanaiah202328} & FM-P2L \cite{wu2023fm} \\
\addlinespace
\midrule
\multirow{12}{*}{2024} & INSPIRE \cite{liu2024inspire} & PACE \cite{wen2024pace} \\
 & PBN \cite{mao2024pbn} & PolyLUT-Add \cite{lou2024polylut} \\
 & SPARK \cite{liu2024spark} & RFMA \cite{lei2024reconfigurable} \\
 & BitWave \cite{shi2024bitwave} & Precision-Scalable \cite{huang2024precision} \\
 & MASL-AFU \cite{meng2024masl} & Trapezoid \cite{yang2024trapezoid} \\
 & AESA \cite{zhang2024efficient} & ViTA \cite{chen2024vita} \\
 & AutoWS \cite{yu2024auto} & Carat \cite{pan2024carat} \\
 & Memory Sharing with CXL \cite{jain2024memory} & Block-Sparsity \cite{lee2024accelerated} \\
 & ENN \cite{jiang2024efficient} & Quantization-aware \cite{chen2024quantization} \\
 & CPoT \cite{geng2024compact} & LUTein \cite{im2024lutein} \\
 & FPMAC \cite{ali2024stochastic} & TreeLUT \cite{khataei2025treelut} \\
 & H2PIPE \cite{doumet2024h2pipe} & \\ 
\addlinespace
\midrule
\multirow{5}{*}{2025} & Flex-EGAI \cite{belano2025flexible} & FIGLUT \cite{park2025figlut} \\
 & FlightVGM \cite{liu2025flightvgm} & TENET \cite{huang2025tenet} \\
 & PIM Is All You Need \cite{gu2025pim} & Panacea \cite{kam2025panacea} \\
 & RLUT \cite{cassidy2025reducedlut} & Flex-PE \cite{lokhande2025flex} \\
 & LUTTC \cite{mo2025lut} & \\ 
\bottomrule
\end{tabular}
\caption{List of Papers}
\label{tab:papers_by_year}
\end{table*}

\begin{figure*}[htbp]
    \centering
    \includegraphics[width=0.9\textwidth]
    {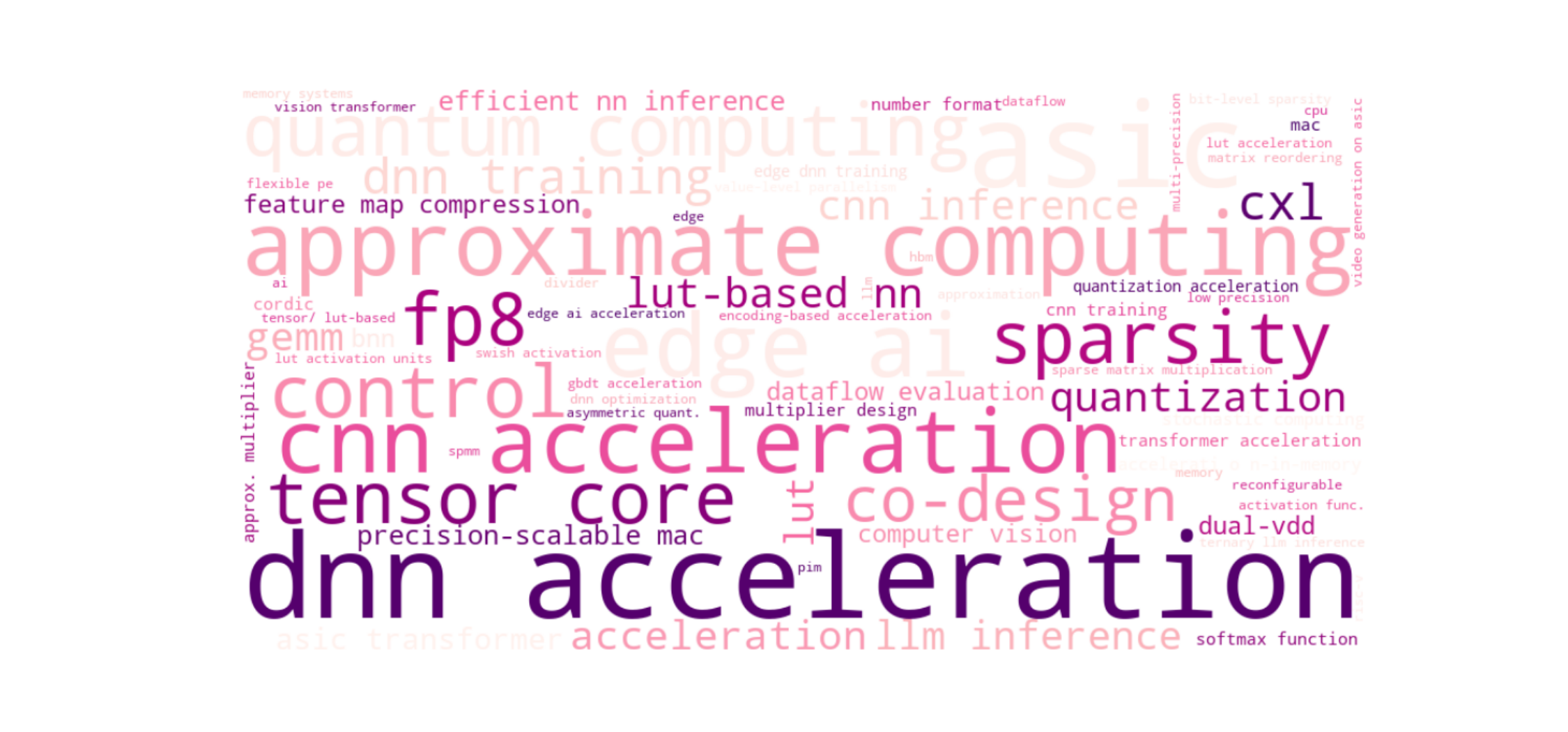}
    \caption{Theme of the papers.}
    \label{fig:keyword_wordcloud}
\end{figure*}

\section{Code-Level Evaluation Scores}
\label{sec:code_level_detailed_scores}
In Table \ref{table:combined_evaluation_scores}, five key metrics are folded into RTL and TB, and we record detailed scores in all dimensions for code-level in Table \ref{table:code_evaluation_scores}.


\begin{table*}[t]
\centering
\footnotesize
    \begin{tabular}{llccccc} 
    \toprule
    \multirow{2}{*}{Method} & \multirow{2}{*}{LLM} & \multicolumn{2}{c}{RTL} & \multicolumn{3}{c}{TB} \\
    \cmidrule(lr){3-4} \cmidrule(lr){5-7}
    & & \multicolumn{1}{c}{Robust.} & \multicolumn{1}{c}{Read.} & \multicolumn{1}{c}{Cov.} & \multicolumn{1}{c}{Assert.} & \multicolumn{1}{c}{Stim.} \\
    \midrule
    Direct & Gemini 2.0 Flash & 30.33 & 43.33 & 23.61 & 19.11 & 32.61 \\
    Direct & Qwen3-Coder-480B & 42.08 & 50.83 & 46.74 & 41.74 & 42.99 \\
    Direct & GPT-4o & 32.50 & 49.72 & 35.27 & 23.61 & 34.72 \\
    Direct & o3-mini & 52.33 & 62.33 & 53.71 & 31.11 & 45.11 \\
    \midrule
    ChatDev & GPT-4o & 17.08 & 17.92 & 17.63 & 1.11 & 11.03 \\
    VerilogCoder & o3-mini & 34.62 & 52.50 & 00.00 & 00.00 & 00.00 \\
    PaperCoder & o3-mini & 51.07 & 49.76 & 45.20 & 20.81 & 35.78 \\
    \midrule
    ArchCraft & Gemini 2.0 Flash & 53.38 & 61.73 & 61.16 & 36.73 & 58.30 \\
    ArchCraft & Qwen3-Coder-480B & 62.91 & 70.51 & 72.25 & \textbf{64.44} & 68.51 \\
    ArchCraft & GPT-4o & 68.25 & 77.03 & 69.50 & 60.93 & 68.37 \\
    ArchCraft & o3-mini & \textbf{80.95} & \textbf{91.13} & \textbf{75.81} & 62.58 & \textbf{74.77} \\
    \bottomrule
    \end{tabular}

\caption{The code-level evaluation is based on five key metrics. The best performance is denoted in \textbf{bold}.}
\label{table:code_evaluation_scores}
\end{table*}

\section{Original Scores}
\label{sec:original_scores}
In this section, we provide the raw score lists from both machine and human evaluations as illustrative examples. Given that an exhaustive presentation of all circuit modules is infeasible, we selectively present the scores for nearly 100 circuit modules derived from the 10 papers in batch1 of ArchSynthBench, as evaluated by othermethod-o3-mini and archcraft-o3-mini.

Specifically, we provide the LLM evaluation results of Direct-o3-mini from Table \ref{table:combined_evaluation_scores}, presented in Table \ref{table:baseline_o3_results_machine}, and the human evaluation results from Table \ref{table:ablation}, presented in Table \ref{table:baseline_o3_results_human}; we also provide the LLM evaluation results of ArchCraft-o3-mini from Table \ref{table:combined_evaluation_scores}, presented in Tables \ref{table:archcraft_o3_results_machine_1} and \ref{table:archcraft_o3_results_machine_2}, as well as the human evaluation results from Table \ref{table:ablation}, presented in Table \ref{table:archcraft_o3_agent_results_human_adjusted_v2}.

\begin{table*}[htbp]
\centering
\setlength{\tabcolsep}{3pt}
\begin{tabular}{lcccccccc}
\toprule
\midrule
\multirow{2}{*}{Design Name} & \multicolumn{3}{c}{Paper-level} & \multicolumn{5}{c}{Code-level} \\
\cmidrule(lr){2-4} \cmidrule(lr){5-9}
& \multicolumn{1}{c}{Fidelity} & \multicolumn{1}{c}{Structural} & \multicolumn{1}{c}{Synth.} & \multicolumn{2}{c}{RTL} & \multicolumn{3}{c}{TB} \\
\cmidrule(lr){5-6} \cmidrule(lr){7-9}
& & & & \multicolumn{1}{c}{Sou.} & \multicolumn{1}{c}{Mai.} & \multicolumn{1}{c}{TC} & \multicolumn{1}{c}{AQ} & \multicolumn{1}{c}{SQ} \\
\midrule
Flex-EGAI & 4.0 & 2.0 & 5.0 & 5.0 & 6.0 & 5.0 & 3.0 & 4.0 \\
OBS-TA & 4.0 & 4.0 & 3.5 & 6.0 & 6.5 & 4.6 & 5.0 & 4.5 \\
INSPIRE  & 4.0 & 4.0 & 4.0 & 5.5 & 6.5 & 5.5 & 2.0 & 4.5 \\
LC-MAC  & 4.0 & 4.5 & 4.0 & 5.5 & 6.5 & 5.0 & 2.0 & 4.0 \\
PBN  & 3.0 & 3.5 & 2.5 & 4.5 & 6.0 & 4.5 & 0.0 & 2.0 \\
SPARK & 2.0 & 3.0 & 4.0 & 5.0 & 6.0 & 5.0 & 3.0 & 5.0 \\
BitWave & 3.0 & 4.0 & 3.5 & 4.5 & 6.0 & 5.0 & 3.0 & 4.5 \\
FlightVGM & 1.0 & 2.0 & 3.0 & 4.5 & 6.0 & 6.0 & 5.5 & 5.0 \\
MASL-AFU & 3.0 & 3.0 & 3.0 & 5.5 & 6.0 & 6.0 & 2.0 & 4.5 \\
ST-Purning & 3.0 & 3.5 & 3.0 & 5.5 & 6.0 & 6.0 & 4.5 & 6.0 \\
\midrule
\bottomrule
\end{tabular}
\caption{Machine Scoring Results Using o3-mini as the LLM  in the Direct Method.}
\label{table:baseline_o3_results_machine}
\end{table*}

\begin{table*}[htbp]
\centering
\setlength{\tabcolsep}{3pt}
\begin{tabular}{lcccccccc}
\toprule
\midrule
\multirow{2}{*}{Paper Name} & \multicolumn{3}{c}{Paper-level} & \multicolumn{5}{c}{Code-level} \\
\cmidrule(lr){2-4} \cmidrule(lr){5-9}
& \multicolumn{1}{c}{Fidelity} & \multicolumn{1}{c}{Structural} & \multicolumn{1}{c}{Synth.} & \multicolumn{2}{c}{RTL} & \multicolumn{3}{c}{TB} \\
\cmidrule(lr){5-6} \cmidrule(lr){7-9}
& & & & \multicolumn{1}{c}{Sou.} & \multicolumn{1}{c}{Mai.} & \multicolumn{1}{c}{TC} & \multicolumn{1}{c}{AQ} & \multicolumn{1}{c}{SQ} \\
\midrule
INSPIRE  & 2.0 & 3.0 & 9.0 & 5.0 & 6.0 & 5.0 & 2.0 & 4.0  \\
MASL-AFU & 3.0 & 2.0 & 8.0 & 5.0 & 6.0 & 6.0 & 2.0 & 4.0  \\
SPARK & 2.0 & 2.0 & 9.0 & 5.0 & 5.0 & 4.0 & 3.0 & 5.0  \\
\midrule
\bottomrule
\end{tabular}
\caption{Human Expert Scoring Results Using o3-mini as the LLM  in the Direct Method.}
\label{table:baseline_o3_results_human}
\end{table*}

\begin{table*}[htbp]
\centering
\setlength{\tabcolsep}{3pt} 
\begin{tabular}{llcccccccc} 
\toprule
\midrule
\multirow{2}{*}{Paper Name} & \multirow{2}{*}{Design Name} & \multicolumn{3}{c}{Paper-level} & \multicolumn{5}{c}{Code-level} \\
\cmidrule(lr){3-5} \cmidrule(lr){6-10}
& & \multicolumn{1}{c}{Fidelity} & \multicolumn{1}{c}{Structural} & \multicolumn{1}{c}{Synth.} & \multicolumn{2}{c}{RTL} & \multicolumn{3}{c}{TB} \\
\cmidrule(lr){6-7} \cmidrule(lr){8-10}
& & & & & \multicolumn{1}{c}{Sou.} & \multicolumn{1}{c}{Mai.} & \multicolumn{1}{c}{TC} & \multicolumn{1}{c}{AQ} & \multicolumn{1}{c}{SQ} \\
\midrule
\multirow{7}{*}{Flex-EGAI} & softex\_streamer & 7.0 & 8.0 & 8.0 & 9.0 & 8.0 & 8.0 & 5.0 & 8.0 \\
& top & 5.0 & 6.0 & 7.0 & 5.0 & 8.0 & 5.0 & 2.0 & 5.0 \\
& tpu & 9.0 & 8.0 & 8.0 & 8.0 & 9.0 & 7.0 & 6.0 & 7.0 \\
& riscv\_cluster & 9.0 & 9.0 & 9.0 & 8.0 & 9.0 & 7.0 & 5.0 & 7.0 \\
& softex\_datapath & 9.0 & 9.0 & 8.0 & 5.0 & 6.0 & 6.0 & 5.0 & 5.0 \\
& softex\_control & 9.0 & 8.0 & 9.0 & 8.0 & 9.0 & 8.0 & 6.0 & 8.0 \\
& memory\_controller & 8.0 & 8.0 & 8.0 & 8.0 & 9.0 & 7.0 & 5.0 & 7.0 \\
\midrule
\multirow{6}{*}{OBS-TA} & config & 10.0 & 9.0 & 9.0 & 10.0 & 10.0 & 8.0 & 8.0 & 8.0 \\
& obs\_datapath & 8.0 & 8.5 & 8.0 & 7.5 & 9.0 & 7.5 & 5.0 & 7.5 \\
& top & 6.0 & 7.5 & 8.0 & 7.0 & 7.5 & 8.0 & 7.0 & 7.5 \\
& obs\_controller & 9.0 & 10.0 & 9.5 & 8.5 & 10.0 & 7.5 & 6.5 & 7.5 \\
& nonlinear\_block & 6.0 & 6.0 & 7.5 & 5.0 & 7.5 & 6.0 & 5.0 & 6.0 \\
& mem\_controller & 9.0 & 8.5 & 9.0 & 4.5 & 7.0 & 6.0 & 4.5 & 6.0 \\
\midrule
\multirow{8}{*}{INSPIRE} & accumulation\_unit & 9.0 & 9.0 & 9.0 & 8.5 & 9.5 & 8.0 & 5.0 & 8.0 \\
& activation\_encoder & 9.0 & 9.0 & 9.0 & 8.5 & 9.0 & 9.0 & 8.0 & 7.5 \\
& control\_unit & 9.0 & 9.5 & 9.0 & 5.5 & 6.5 & 4.5 & 3.0 & 5.5 \\
& ip\_pe & 9.0 & 9.5 & 9.0 & 9.0 & 9.5 & 10.0 & 9.0 & 9.5 \\
& ip\_pe\_array & 9.5 & 9.5 & 10.0 & 9.0 & 9.5 & 8.0 & 7.0 & 7.5 \\
& memory\_interface & 9.0 & 9.0 & 9.0 & 8.0 & 9.0 & 8.0 & 7.0 & 8.0 \\
& top & 8.0 & 9.0 & 8.5 & 7.5 & 9.0 & 7.0 & 5.0 & 7.5 \\
& weight\_buffer & 9.0 & 9.0 & 10.0 & 5.5 & 6.0 & 4.5 & 5.0 & 4.5 \\
\midrule
\multirow{6}{*}{LC-MAC} & bit\_brick & 8.0 & 9.0 & 8.5 & 9.0 & 9.0 & 9.0 & 8.5 & 8.0 \\
& rfu\_core & 7.5 & 9.0 & 8.5 & 8.0 & 8.0 & 8.5 & 6.0 & 7.5 \\
& shift\_add\_unit & 9.0 & 9.0 & 9.0 & 9.0 & 9.5 & 7.5 & 8.0 & 7.5 \\
& bbu\_unit & 9.5 & 9.5 & 9.0 & 9.0 & 9.5 & 7.5 & 5.0 & 7.5 \\
& top & 9.0 & 9.0 & 9.0 & 8.0 & 9.5 & 9.0 & 6.0 & 8.5 \\
& mode\_controller & 9.0 & 9.5 & 10.0 & 9.5 & 9.5 & 8.0 & 7.5 & 8.0 \\
\midrule
\multirow{6}{*}{PBN} & pbn\_control\_unit & 9.0 & 9.0 & 9.0 & 8.5 & 9.0 & 8.5 & 5.0 & 7.5 \\
& pbn\_datapath & 8.0 & 7.0 & 7.5 & 7.5 & 9.0 & 8.0 & 5.0 & 7.5 \\
& pbn\_fifo & 10.0 & 9.0 & 10.0 & 9.0 & 9.5 & 8.5 & 8.0 & 9.0 \\
& pbn\_nps & 8.0 & 9.0 & 8.0 & 7.5 & 9.0 & 7.0 & 5.0 & 7.0 \\
& pbn\_scs & 9.0 & 9.0 & 8.5 & 8.0 & 9.0 & 7.0 & 5.0 & 7.0 \\
& pbn\_top & 9.0 & 10.0 & 9.0 & 8.5 & 10.0 & 7.0 & 6.0 & 7.5 \\
\midrule
\bottomrule
\end{tabular}
\caption{Machine Scoring Results Using o3-mini as the LLM  in the ArchCraft. I }
\label{table:archcraft_o3_results_machine_1}
\end{table*}

\begin{table*}[htbp]
\centering
\setlength{\tabcolsep}{3pt} 
\begin{tabular}{llcccccccc} 
\toprule
\midrule
\multirow{2}{*}{Paper Name} & \multirow{2}{*}{Design Name} & \multicolumn{3}{c}{Paper-level} & \multicolumn{5}{c}{Code-level} \\
\cmidrule(lr){3-5} \cmidrule(lr){6-10}
& & \multicolumn{1}{c}{Fidelity} & \multicolumn{1}{c}{Structural} & \multicolumn{1}{c}{Synth.} & \multicolumn{2}{c}{RTL} & \multicolumn{3}{c}{TB} \\
\cmidrule(lr){6-7} \cmidrule(lr){8-10}
& & & & & \multicolumn{1}{c}{Sou.} & \multicolumn{1}{c}{Mai.} & \multicolumn{1}{c}{TC} & \multicolumn{1}{c}{AQ} & \multicolumn{1}{c}{SQ} \\
\midrule
\multirow{8}{*}{SPARK} & control\_unit & 9.0 & 9.0 & 9.0 & 9.0 & 10.0 & 8.0 & 9.0 & 8.0 \\
& im2col\_pack\_engine & 9.0 & 9.5 & 10.0 & 9.0 & 9.5 & 7.5 & 5.0 & 8.5 \\
& interface & 8.5 & 8.0 & 9.0 & 8.0 & 8.0 & 6.0 & 6.0 & 7.0 \\
& memory\_controller & 9.0 & 9.0 & 9.5 & 8.5 & 9.0 & 7.0 & 6.0 & 7.0 \\
& pe\_array & 8.0 & 9.0 & 9.0 & 9.0 & 9.5 & 7.0 & 5.0 & 7.0 \\
& spark\_decoder & 10.0 & 10.0 & 10.0 & 10.0 & 10.0 & 9.0 & 5.0 & 7.5 \\
& spark\_encoder & 9.0 & 9.0 & 9.5 & 9.5 & 10.0 & 8.0 & 7.0 & 7.0 \\
& top & 9.0 & 9.0 & 9.0 & 8.0 & 9.0 & 7.5 & 7.0 & 7.0 \\
\midrule
\multirow{7}{*}{BitWave} & bce\_array & 9.0 & 9.0 & 8.5 & 8.5 & 9.5 & 9.0 & 8.0 & 9.0 \\
& control\_unit & 9.0 & 8.5 & 9.0 & 8.0 & 9.0 & 8.0 & 7.5 & 8.0 \\
& data\_dispatcher & 9.0 & 8.0 & 9.0 & 8.0 & 9.0 & 7.0 & 6.0 & 7.0 \\
& memory\_interface & 8.0 & 9.0 & 9.0 & 7.5 & 9.0 & 7.0 & 6.0 & 7.0 \\
& output\_formatter & 10.0 & 10.0 & 9.5 & 9.5 & 9.0 & 9.0 & 7.5 & 8.5 \\
& top & 8.0 & 9.0 & 9.0 & 7.5 & 8.5 & 8.0 & 5.0 & 8.0 \\
& zcip\_parser & 9.5 & 9.5 & 10.0 & 9.0 & 9.0 & 8.0 & 7.0 & 8.0 \\
\midrule
\multirow{10}{*}{FlightVGM} & matrix\_processing\_engine & 6.0 & 6.0 & 7.0 & 5.0 & 7.0 & 6.0 & 4.0 & 5.0 \\
& cpu\_scheduler & 9.0 & 8.5 & 9.0 & 8.5 & 9.0 & 8.5 & 7.5 & 8.0 \\
& top & 5.0 & 4.0 & 7.5 & 6.0 & 7.0 & 6.0 & 4.0 & 5.0 \\
& recovery\_unit & 9.0 & 8.5 & 9.0 & 9.0 & 10.0 & 8.5 & 8.0 & 8.5 \\
& global\_interconnect & 7.0 & 9.0 & 9.0 & 7.5 & 10.0 & 9.0 & 9.0 & 9.0 \\
& sparsification\_unit & 9.0 & 8.5 & 8.0 & 8.5 & 9.0 & 7.0 & 5.0 & 7.0 \\
& dsp\_e & 8.0 & 8.5 & 8.0 & 7.5 & 9.0 & 6.0 & 6.0 & 7.0 \\
& mmu\_controller & 9.0 & 8.5 & 9.0 & 8.5 & 9.5 & 8.0 & 7.5 & 7.5 \\
& compute\_core & 6.0 & 4.0 & 6.0 & 7.0 & 8.0 & 7.0 & 5.0 & 6.0 \\
& special\_function\_unit & 8.0 & 8.0 & 7.5 & 7.5 & 8.5 & 7.0 & 5.0 & 6.0 \\
\midrule
\multirow{10}{*}{MASL-AFU} & bsearch\_unit & 9.0 & 8.0 & 9.0 & 9.0 & 8.0 & 8.0 & 7.0 & 8.0 \\
& control\_unit & 9.0 & 9.0 & 9.0 & 8.0 & 8.5 & 7.5 & 7.0 & 7.5 \\
& lut\_unit & 9.0 & 9.0 & 9.0 & 8.5 & 9.0 & 8.5 & 8.0 & 9.0 \\
& mac\_unit & 9.0 & 9.0 & 8.0 & 8.5 & 9.5 & 8.5 & 8.0 & 7.5 \\
& masl\_afu\_top & 9.0 & 10.0 & 9.0 & 9.0 & 9.5 & 8.0 & 6.0 & 8.0 \\
& masl\_afu\_unit & 9.0 & 9.0 & 9.0 & 8.0 & 9.5 & 7.5 & 6.0 & 7.5 \\
& memory\_controller & 9.0 & 8.5 & 9.0 & 8.0 & 9.0 & 8.0 & 6.0 & 8.0 \\
& scalability\_node & 9.0 & 8.5 & 9.0 & 9.0 & 9.5 & 9.0 & 8.0 & 9.0 \\
& shared\_buffer & 9.0 & 9.0 & 8.5 & 8.5 & 9.5 & 9.0 & 7.5 & 8.0 \\
& top & 9.0 & 9.5 & 9.0 & 9.0 & 10.0 & 6.5 & 5.0 & 6.0 \\
\midrule
\multirow{7}{*}{ST-Pruning} & accumulator & 9.0 & 9.0 & 9.0 & 9.0 & 10.0 & 9.0 & 8.0 & 8.0 \\
& config\_interface & 8.0 & 9.0 & 9.0 & 8.5 & 9.0 & 7.5 & 5.0 & 7.5 \\
& control\_unit & 9.0 & 9.0 & 9.0 & 8.5 & 9.5 & 8.0 & 5.0 & 7.5 \\
& fsde\_encode & 9.0 & 7.5 & 9.0 & 8.0 & 9.0 & 9.0 & 7.0 & 8.0 \\
& gpe\_array & 9.0 & 8.5 & 8.0 & 5.5 & 6.0 & 4.0 & 4.0 & 4.0 \\
& memory\_controller & 9.0 & 9.0 & 9.0 & 8.5 & 9.0 & 7.5 & 7.5 & 7.5 \\
& top & 8.0 & 9.0 & 8.5 & 7.5 & 9.0 & 7.0 & 6.0 & 7.5 \\
\midrule
\bottomrule
\end{tabular}
\caption{Machine Scoring Results Using o3-mini as the LLM  in the ArchCraft. II }
\label{table:archcraft_o3_results_machine_2}
\end{table*}

\begin{table*}[htbp]
\centering
\setlength{\tabcolsep}{3pt} 
\begin{tabular}{llcccccccc} 
\toprule
\midrule
\multirow{2}{*}{Paper Name} & \multirow{2}{*}{Design Name} & \multicolumn{3}{c}{Paper-level} & \multicolumn{5}{c}{Code-level} \\
\cmidrule(lr){3-5} \cmidrule(lr){6-10}
& & \multicolumn{1}{c}{Fidelity} & \multicolumn{1}{c}{Structural} & \multicolumn{1}{c}{Synth.} & \multicolumn{2}{c}{RTL} & \multicolumn{3}{c}{TB} \\
\cmidrule(lr){6-7} \cmidrule(lr){8-10}
& & & & & \multicolumn{1}{c}{Sou.} & \multicolumn{1}{c}{Mai.} & \multicolumn{1}{c}{TC} & \multicolumn{1}{c}{AQ} & \multicolumn{1}{c}{SQ} \\
\midrule
\multirow{8}{*}{INSPIRE} & accumulation\_unit & 10.00 & 7.33 & 8.50 & 9.17 & 9.00 & 8.33 & 5.67 & 8.33 \\
& activation\_encoder & 9.17 & 7.67 & 8.33 & 9.00 & 9.17 & 9.33 & 8.67 & 7.50 \\
& control\_unit & 5.50 & 8.33 & 5.67 & 5.00 & 5.50 & 4.33 & 3.33 & 5.17 \\
& ip\_pe & 9.17 & 9.67 & 9.00 & 9.33 & 9.17 & 10.00 & 9.33 & 9.67 \\
& ip\_pe\_array & 9.00 & 10.00 & 9.50 & 9.67 & 9.00 & 8.67 & 7.33 & 8.50 \\
& memory\_interface & 7.67 & 8.50 & 9.17 & 8.33 & 8.67 & 7.50 & 6.17 & 7.33 \\
& top & 7.33 & 7.50 & 7.67 & 7.17 & 7.33 & 5.50 & 4.67 & 4.50 \\
& weight\_buffer & 5.17 & 5.33 & 7.50 & 5.67 & 6.17 & 4.67 & 5.50 & 4.17 \\
\midrule
\multirow{8}{*}{SPARK} & control\_unit & 9.00 & 9.50 & 9.17 & 9.33 & 10.00 & 8.67 & 9.33 & 8.50 \\
& im2col\_pack\_engine & 10.00 & 9.33 & 10.00 & 9.17 & 10.00 & 7.33 & 5.17 & 8.67 \\
& interface & 8.33 & 7.50 & 8.17 & 8.00 & 8.33 & 6.67 & 6.50 & 7.17 \\
& memory\_controller & 9.17 & 9.00 & 9.50 & 9.17 & 9.67 & 7.50 & 5.67 & 7.33 \\
& pe\_array & 8.67 & 9.17 & 9.00 & 9.00 & 9.50 & 7.17 & 5.50 & 7.50 \\
& spark\_decoder & 10.00 & 10.00 & 10.00 & 10.00 & 10.00 & 9.33 & 5.67 & 7.67 \\
& spark\_encoder & 10.00 & 10.00 & 10.00 & 10.00 & 10.00 & 8.50 & 7.17 & 7.50 \\
& top & 8.50 & 8.17 & 8.33 & 8.67 & 8.00 & 7.67 & 7.33 & 7.67 \\
\midrule
\multirow{10}{*}{MASL-AFU} & bsearch\_unit & 9.33 & 8.50 & 9.17 & 9.00 & 8.67 & 8.33 & 7.67 & 7.50 \\
& control\_unit & 9.17 & 9.00 & 9.50 & 9.67 & 9.33 & 7.50 & 7.17 & 7.67 \\
& lut\_unit & 7.50 & 7.67 & 8.33 & 8.50 & 8.00 & 8.17 & 8.67 & 8.50 \\
& mac\_unit & 9.00 & 9.50 & 8.17 & 8.33 & 10.00 & 8.67 & 8.50 & 8.33 \\
& masl\_afu\_top & 9.67 & 8.33 & 8.67 & 8.50 & 9.17 & 8.00 & 6.33 & 7.17 \\
& masl\_afu\_unit & 9.50 & 9.17 & 8.50 & 8.00 & 10.00 & 7.67 & 7.50 & 7.33 \\
& memory\_controller & 8.67 & 8.50 & 9.33 & 9.50 & 9.00 & 8.33 & 7.17 & 7.67 \\
& scalability\_node & 9.00 & 7.67 & 10.00 & 10.00 & 10.00 & 9.50 & 8.33 & 9.17 \\
& shared\_buffer & 9.17 & 8.33 & 9.00 & 9.00 & 9.50 & 7.17 & 5.50 & 8.67 \\
& top & 8.50 & 9.00 & 8.67 & 8.17 & 9.00 & 7.50 & 5.67 & 7.17 \\
\midrule
\bottomrule
\end{tabular}
\caption{Overall Human Scoring Results Using o3-mini as the LLM in ArchCraft.}
\label{table:archcraft_o3_agent_results_human_adjusted_v2}
\end{table*}

\section{Analysis}
\label{sec:rectifying_analysis}
\textbf{Rectifying Analysis.}
Our analysis focus on the mean and variance of feedback-based rectification steps needed to correct code errors within the ArchCraft framework, utilizing Gemini 2.0 Flash, Qwen3-Coder-480B, GPT-4o \cite{openai_gpt4o} and o3-mini as backbone models. As Table \ref{table:rectifying_analysis_results} illustrates, Gemini 2.0 Flash required the most rectification iterations, averaging 7.58, indicating a significant need for corrections. Conversely, o3-mini showed the lowest average and minimal variance in rectification counts, suggesting it consistently produced more stable outputs. These findings collectively indicate that our framework not only enhances the performance of all LLM models through its correction process, but also that the RTL code generated by o3-mini possesses superior initial correctness and lower integration costs.

\begin{table}[htbp]
\centering
\resizebox{\columnwidth}{!}{%
\begin{tabular}{lccc}
\toprule
LLM Model & CD Num. & Mean & Variance \\
\midrule
Gemini 2.0 Flash & 92 & 7.58 & 8.52 \\
Qwen3-Coder-480B & 87 & 6.44 & 8.47 \\
GPT-4o & 70 & 4.95 & 9.26 \\
o3-mini & 75 & 2.07 & 7.64 \\
\bottomrule
\end{tabular}
}
\caption{Rectifying phase iteration counts. CD Num. is the abbreviation of Circuit Designs Number}
\label{table:rectifying_analysis_results} 
\end{table}

\section{Physical Implementation}
\label{sec:physical_implementation}
\begin{figure}[htbp]
    \centering
    \includegraphics[width=0.47\textwidth]{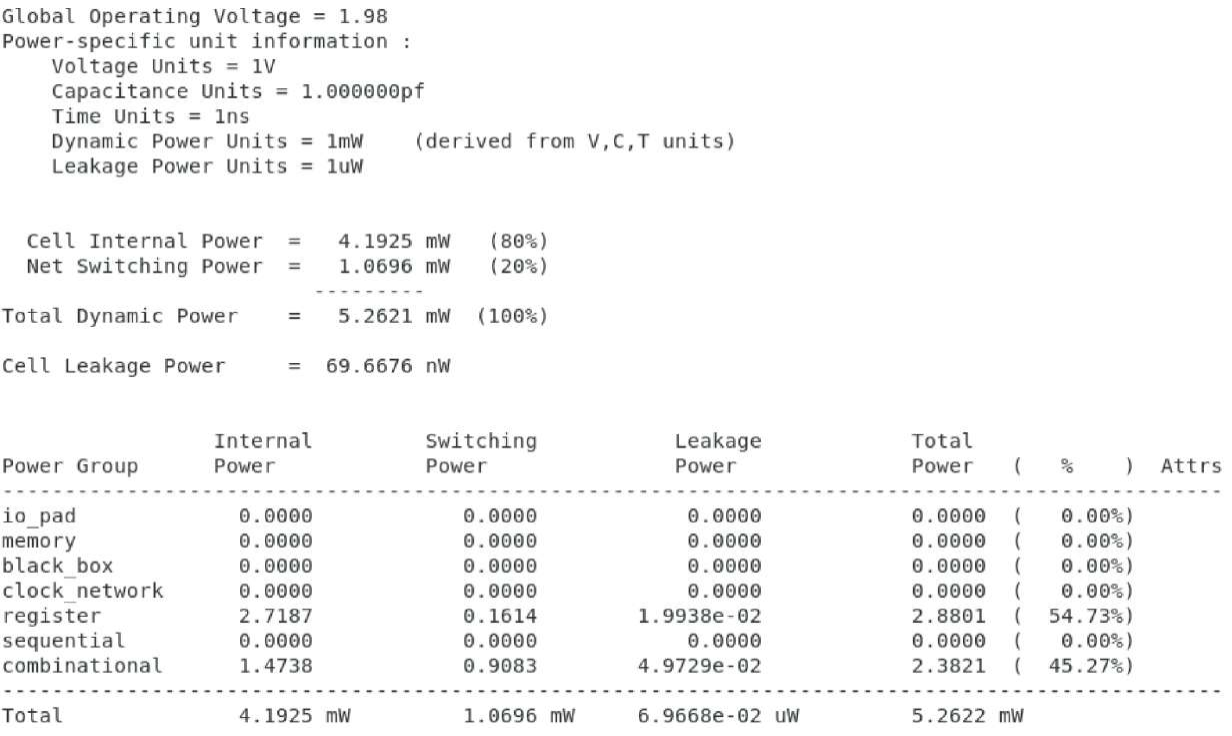}
    \caption{Reported power of the pe\_array.}
    \label{fig:s_p_p}
\end{figure}

\begin{figure}[htbp]
    \centering
    \includegraphics[width=0.47\textwidth]{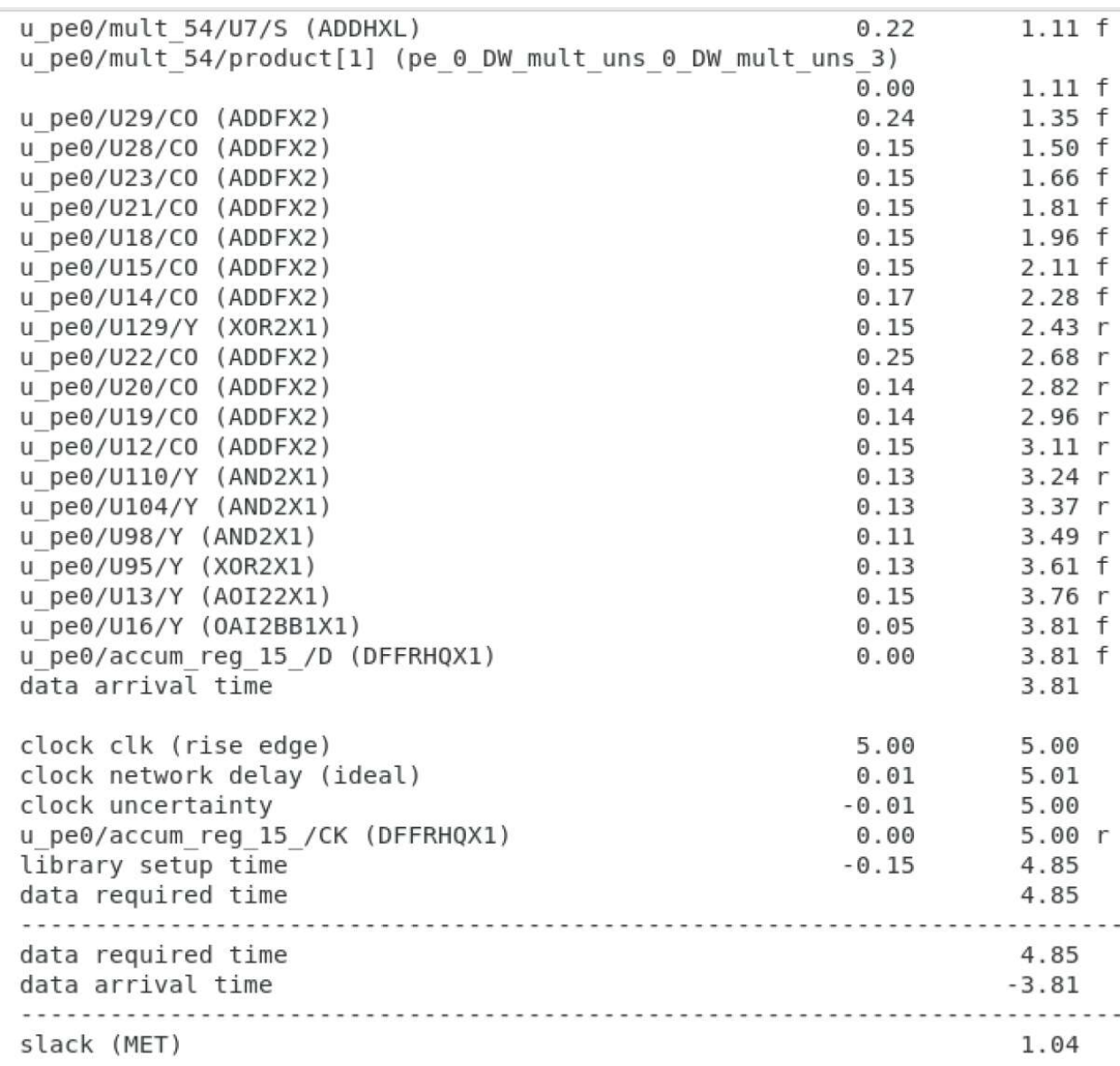}
    \caption{Reported performance of the pe\_array.}
    \label{fig:s_p_t}
\end{figure}

\begin{figure}[htbp]
    \centering
    \includegraphics[width=0.47\textwidth]{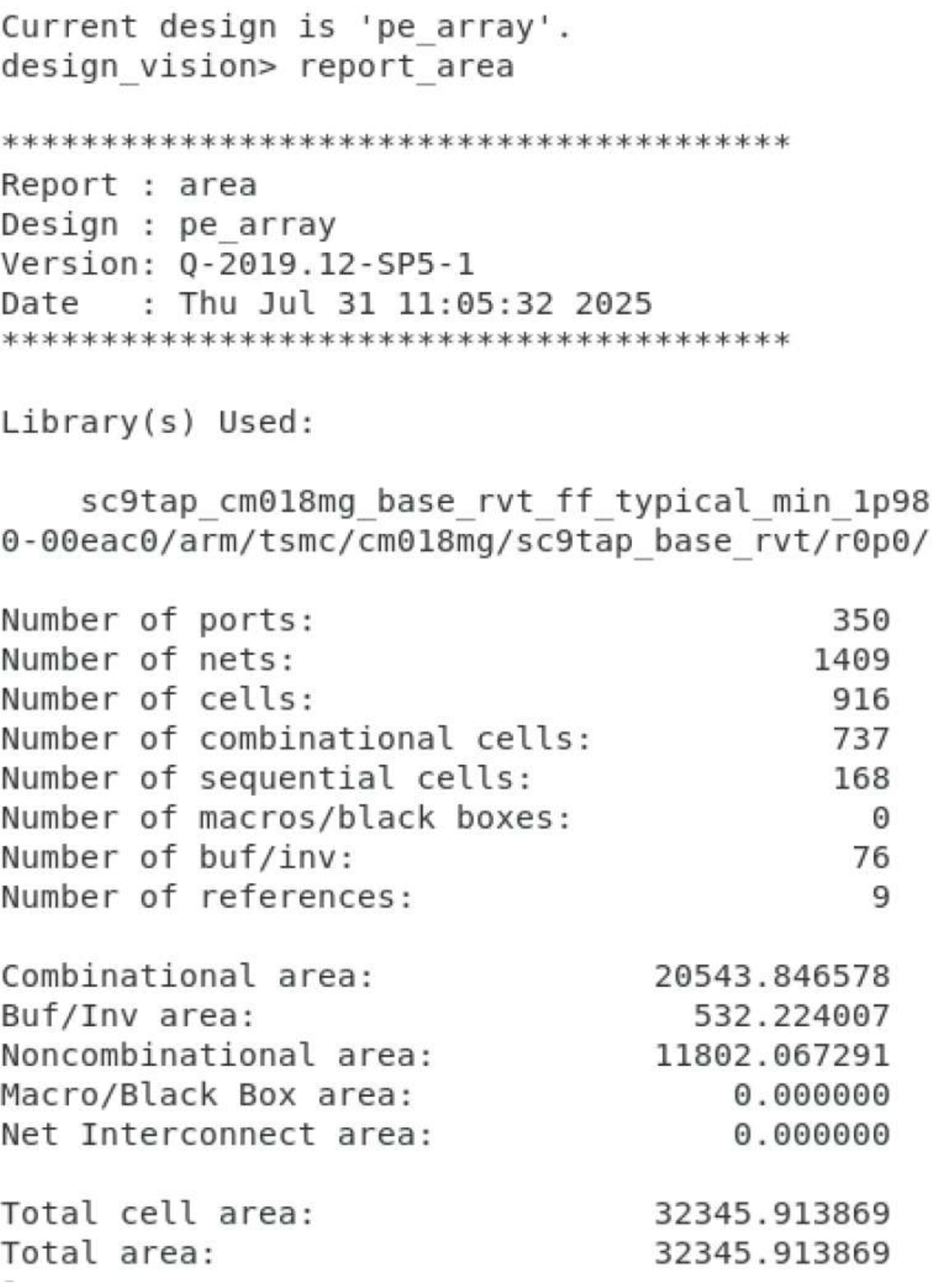}
    \caption{Reported area of the pe\_array.}
    \label{fig:s_p_a}
\end{figure}

\begin{figure}[htbp]
    \centering
    \includegraphics[width=0.47\textwidth]{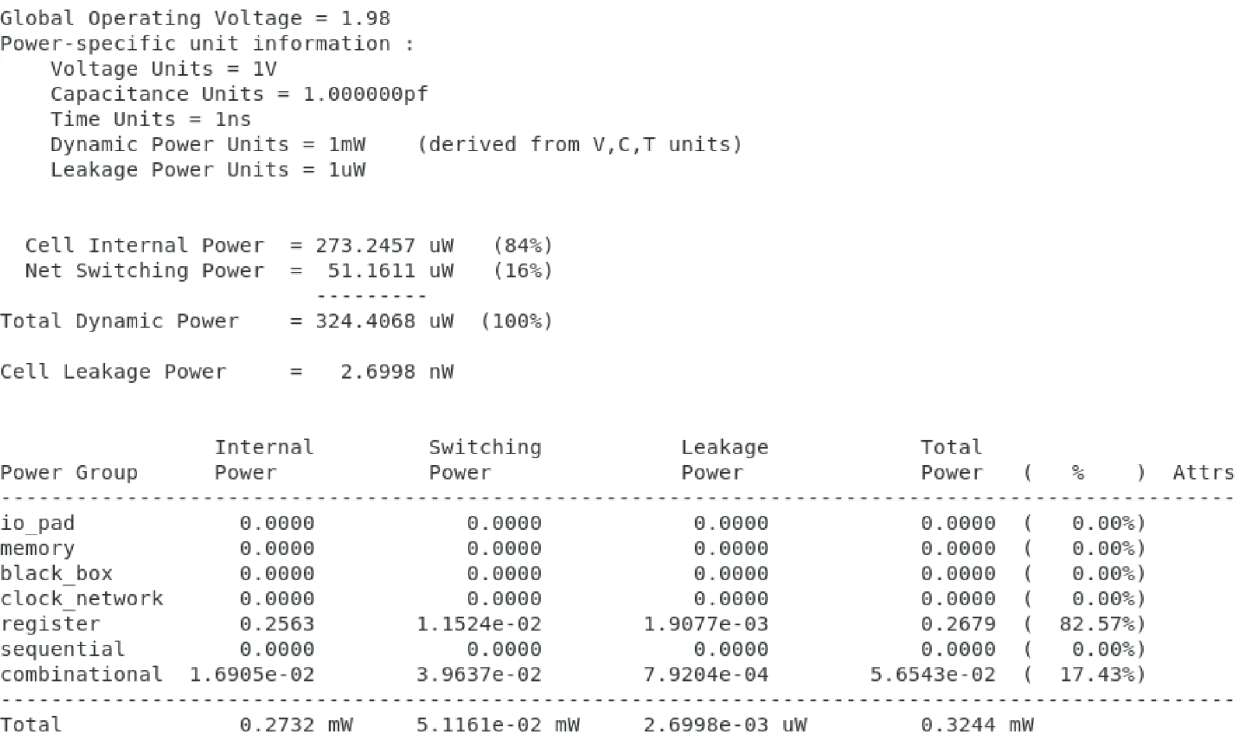}
    \caption{Reported power of the spark\_encoder.}
    \label{fig:s_e_p}
\end{figure}

\begin{figure}[htbp]
    \centering
    \includegraphics[width=0.47\textwidth]{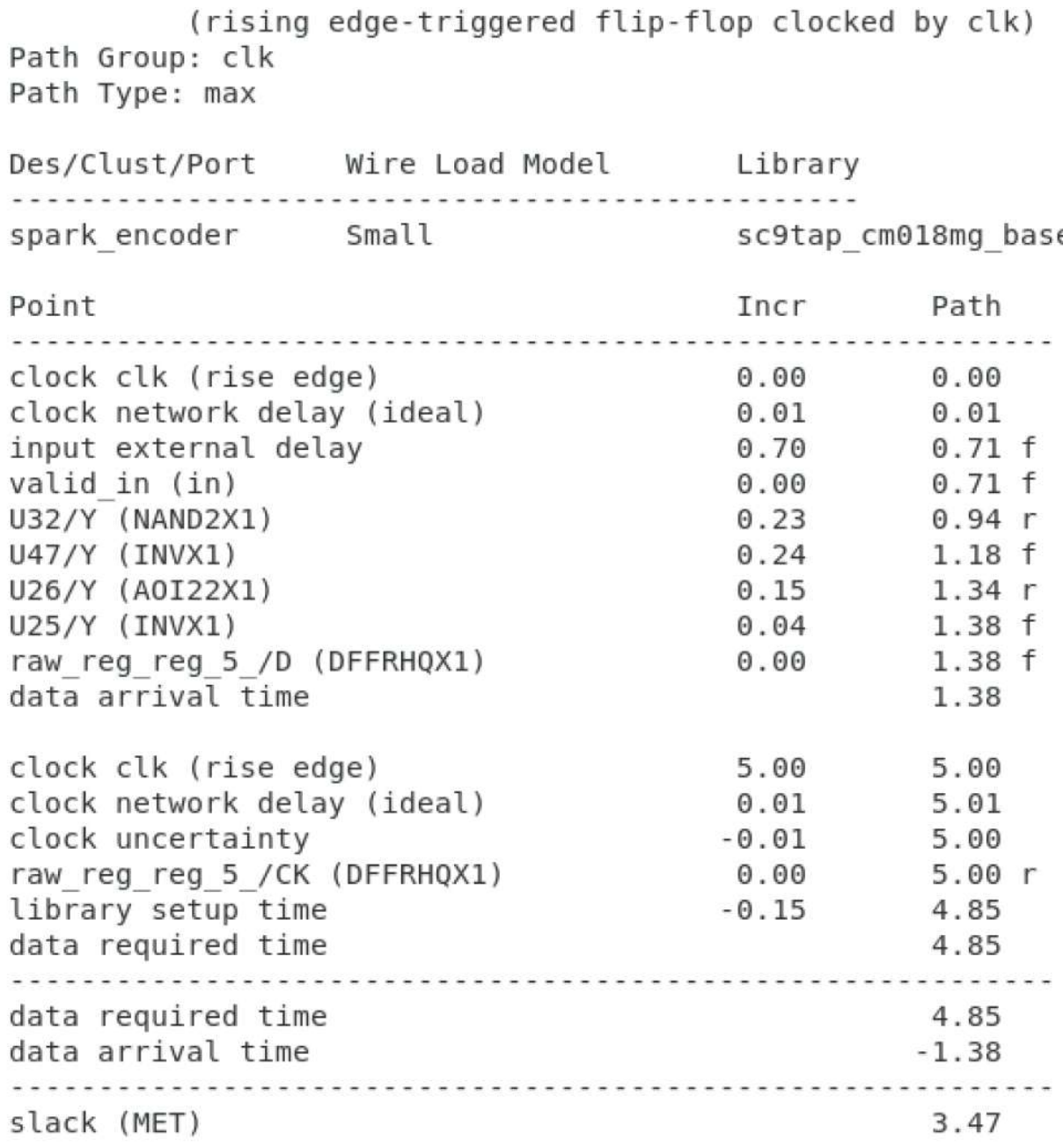}
    \caption{Reported performance of the spark\_encoder.}
    \label{fig:s_e_t}
\end{figure}

\begin{figure}[htbp]
    \centering
    \includegraphics[width=0.47\textwidth]{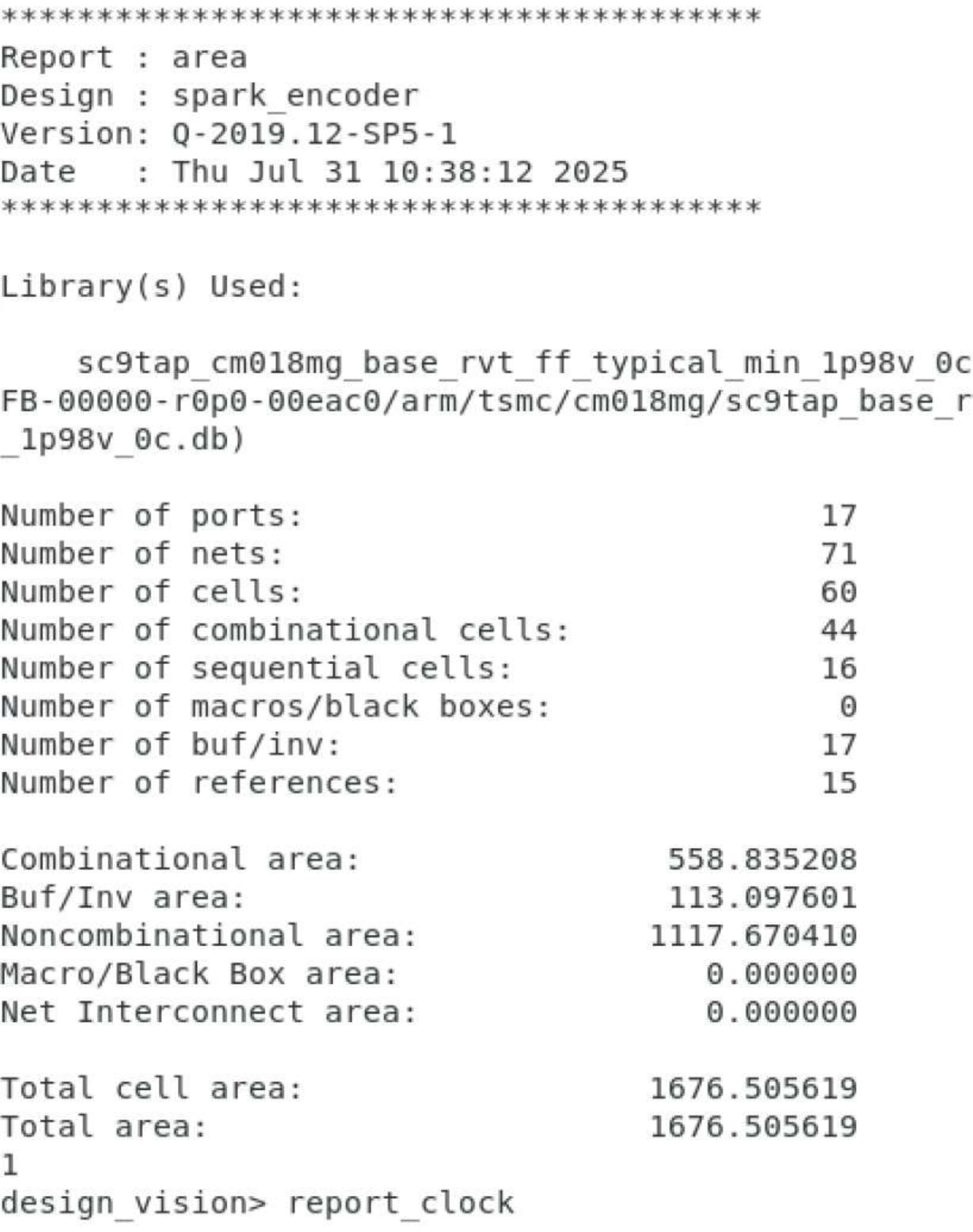}
    \caption{Reported area of the spark\_encoder.}
    \label{fig:s_e_a}
\end{figure}

\begin{figure}[htbp]
    \centering
    \includegraphics[width=0.47\textwidth]{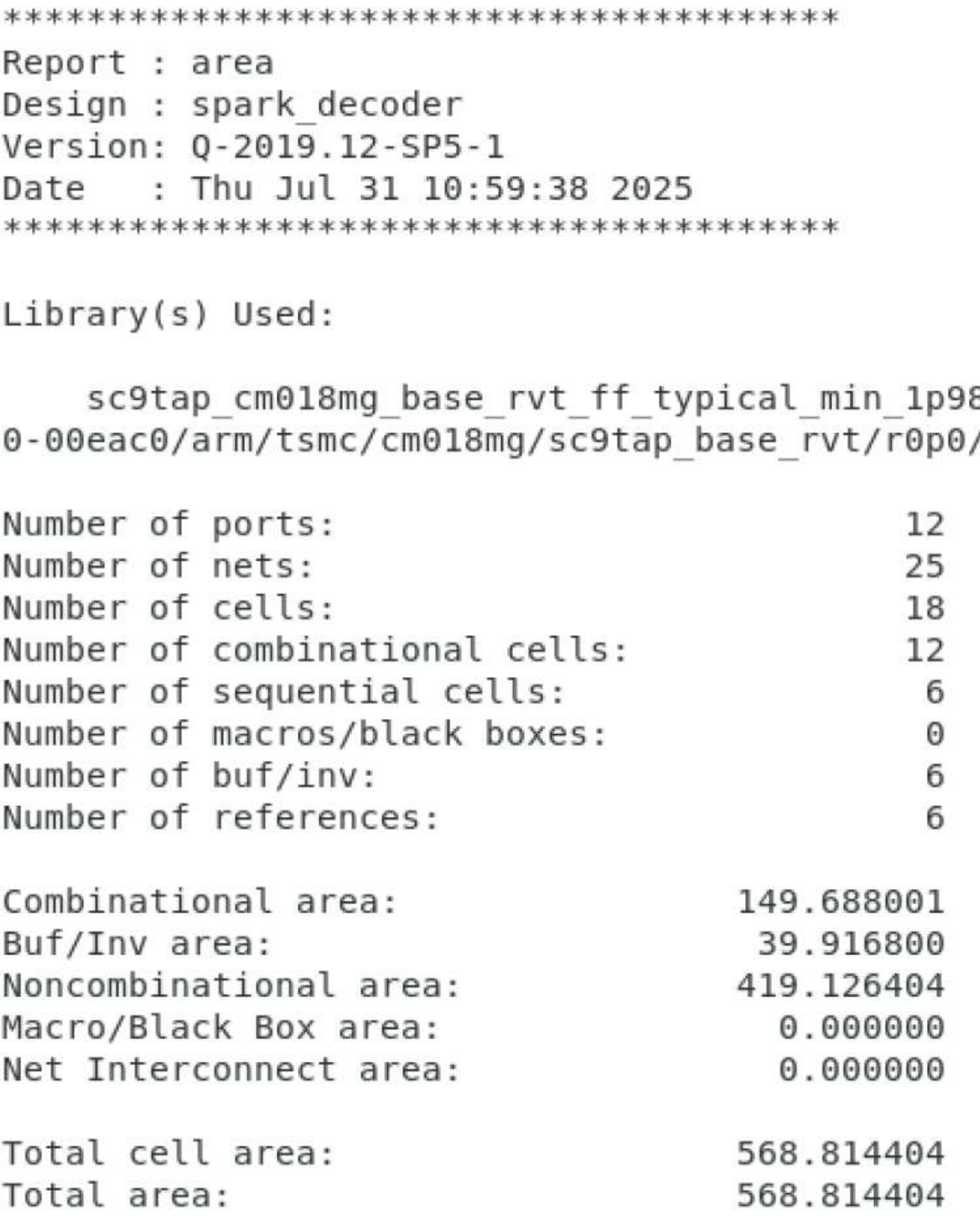}
    \caption{Reported power of the spark\_decoder.}
    \label{fig:s_d_p}
\end{figure}

\begin{figure}[htbp]
    \centering
    \includegraphics[width=0.47\textwidth]{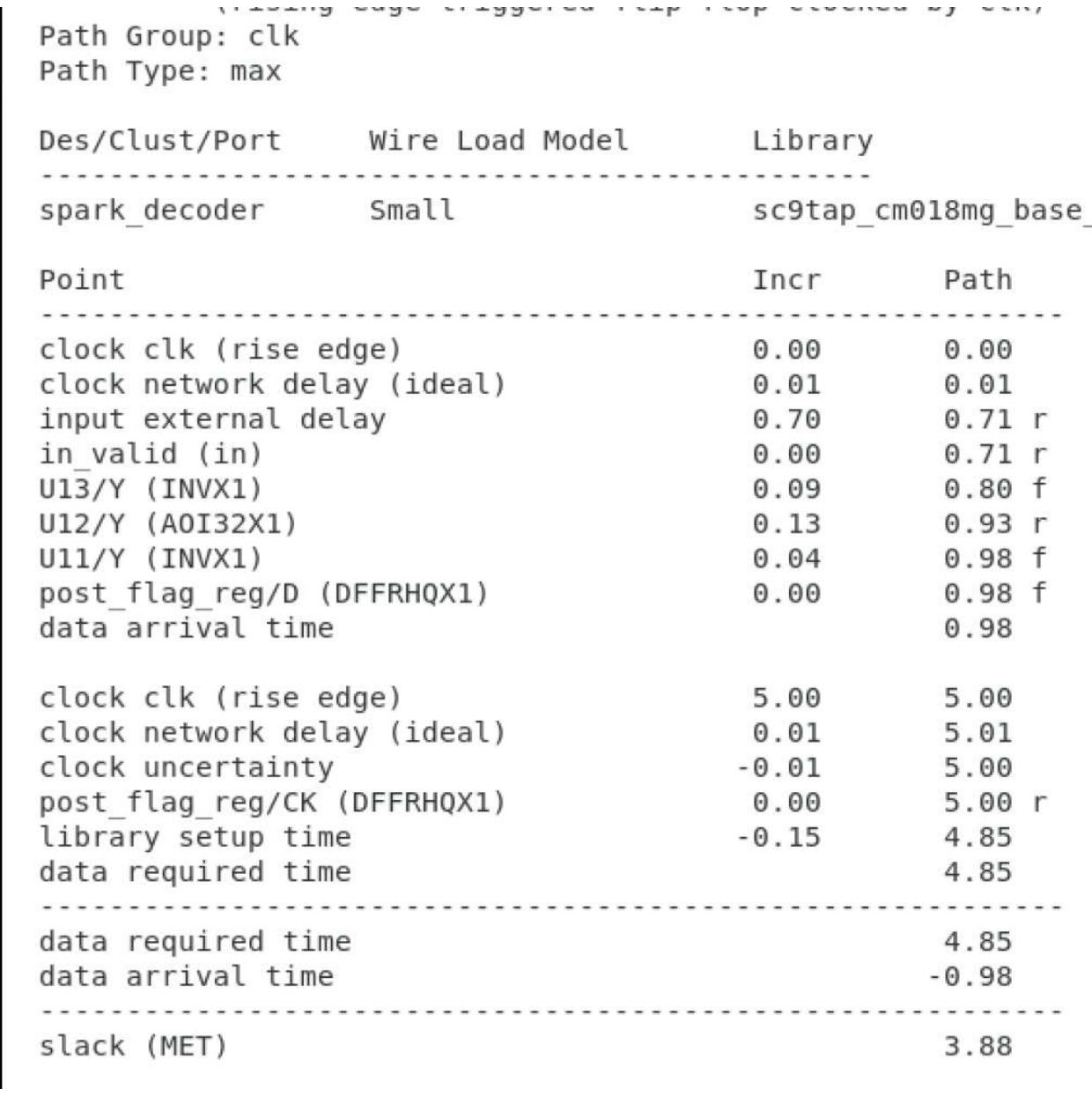}
    \caption{Reported performance of the spark\_decoder.}
    \label{fig:s_d_t}
\end{figure}

\begin{figure}[htbp]
    \centering
    \includegraphics[width=0.47\textwidth]{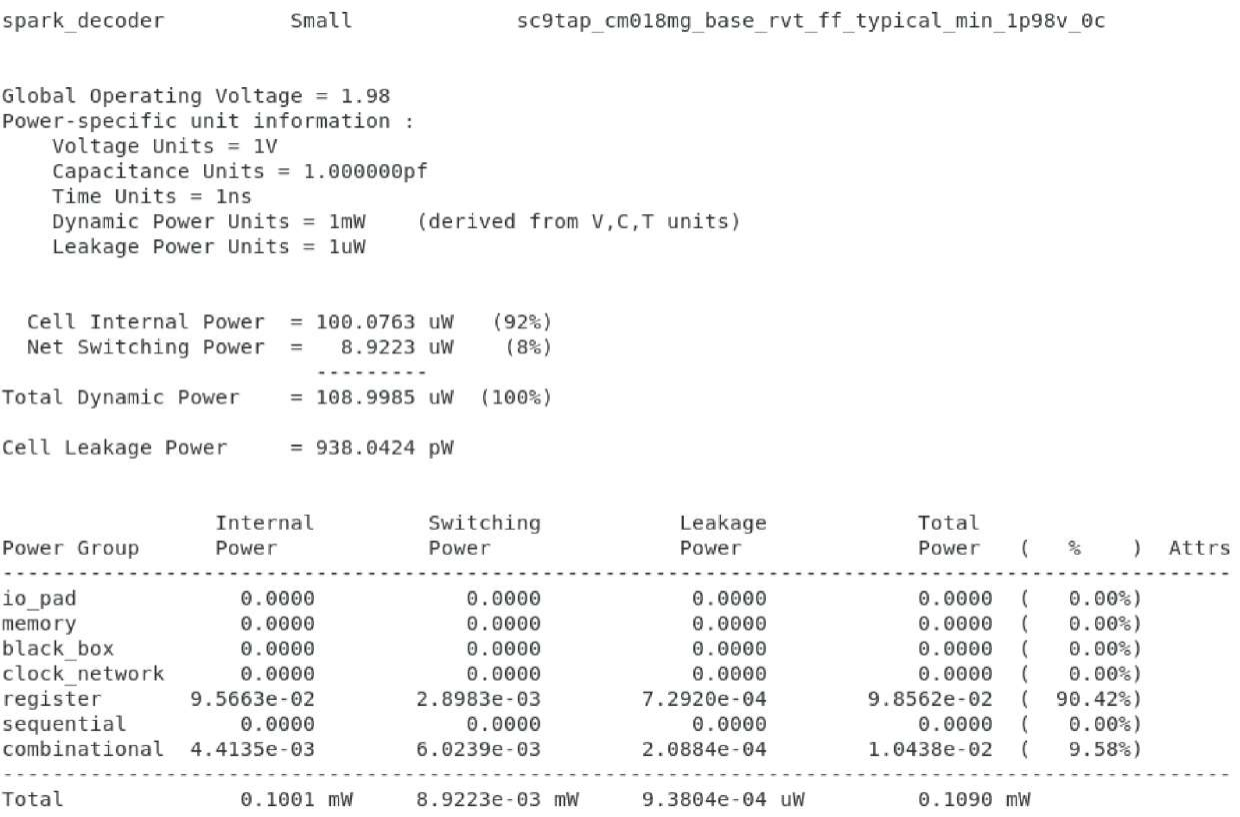}
    \caption{Reported area of the spark\_decoder.}
    \label{fig:s_d_a}
\end{figure}

In this section, we show the DC results of the final case study in the main text, three designs are shown. Figures \ref{fig:s_p_p}, Figure \ref{fig:s_p_t},  Figure \ref{fig:s_p_a} show the PPA of the pe\_array, Figure \ref{fig:s_e_p}, Figure \ref{fig:s_e_t}, Figure \ref{fig:s_e_a} show the PPA of the spark\_encoder, Figure \ref{fig:s_d_p}, Figure \ref{fig:s_d_t}, Figure \ref{fig:s_d_a} show the PPA of the spark\_decoder.

\section{Pearson correlation}
\label{sec:Pearson_correlation}
\begin{figure}[t]
    \centering
    \includegraphics[width=0.49\textwidth]{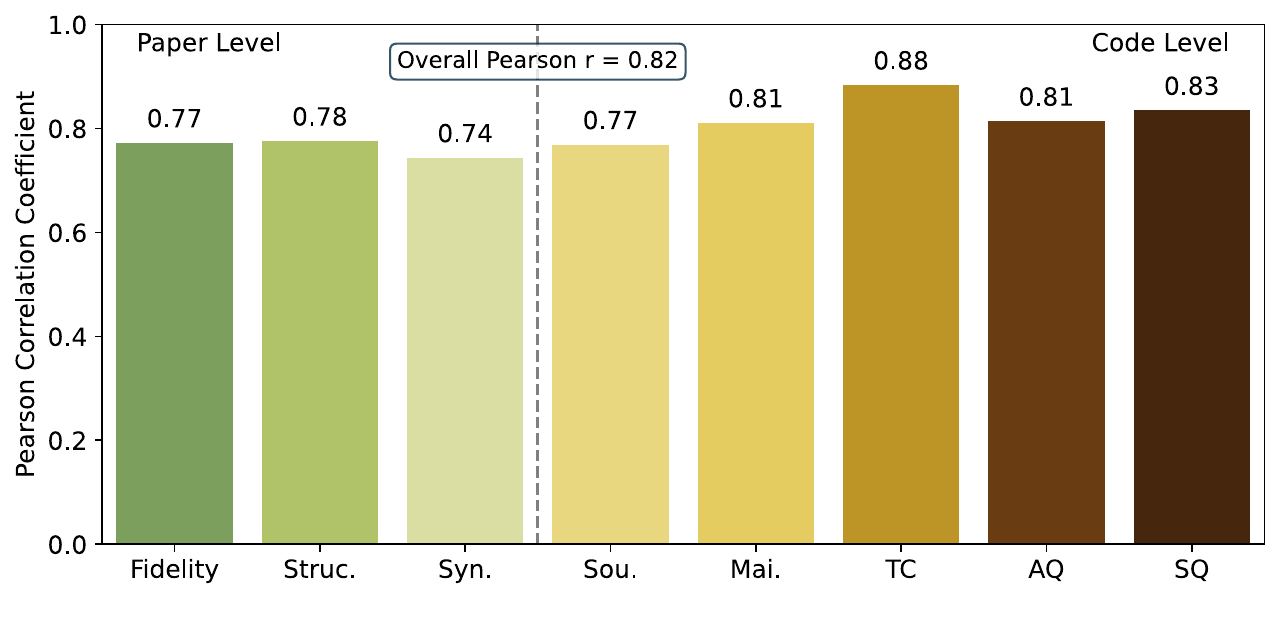}
    \caption{Pearson correlation between human and machine-based scores for ArchCraft}
    \label{fig:Pearson}
\end{figure}

Furthermore, to ascertain the feasibility and reliability of our automated scoring system, we examine their agreement by calculating the Pearson correlation coefficient between the scores assigned by human experts and those derived from our machine-based evaluations, as illustrated in Figure \ref{fig:Pearson}, total Pearson r = 0.82, which demonstrates strong consistency. 

\clearpage



\end{document}